\documentclass[acmsmall,screen]{acmart}
\usepackage[caption=false]{subfig}
\usepackage{multirow}
\usepackage{multicol}
\usepackage{makecell}
\usepackage{threeparttable}
\newcommand{\warning}[1]{\textcolor[RGB]{0, 0, 0}{#1}}

\AtBeginDocument{%
  }

\setcopyright{none}

\begin{document}

\title{Sensing, Social, and Motion Intelligence in Embodied Navigation: A Comprehensive Survey}

\author{Chaoran Xiong}
\email{sjtu4742986@sjtu.edu.cn}
\affiliation{%
  \institution{Shanghai Jiao Tong University}
  \city{Shanghai}
  \country{China}
}

\author{Yulong Huang}
\email{heuedu@163.com}
\affiliation{%
  \institution{Harbin Engineering University}
  \city{Harbin}
  \country{China}
}

\author{Fangwen Yu}
\email{yufangwen@tsinghua.edu.cn}
\affiliation{%
  \institution{Tsinghua University}
  \city{Beijing}
  \country{China}
}

\author{Changhao Chen}
\email{changhaochen@hkust-gz.edu.cn}
\affiliation{%
  \institution{The Hong Kong University of Science and Technology (Guangzhou)}
  \city{Guangzhou}
  \country{China}
}

\author{Yue Wang}
\email{wangyue@iipc.zju.edu.cn}
\affiliation{%
  \institution{Zhejiang University}
  \city{Hangzhou}
  \country{China}
}

\author{Songpengchen Xia}
\email{songpengchengxia@sjtu.edu.cn}
\affiliation{%
  \institution{Shanghai Jiao Tong University}
  \city{Shanghai}
  \country{China}
}

\author{Ling Pei}
\authornote{Corresponding author: Ling Pei.}
\email{ling.pei@sjtu.edu.cn}
\affiliation{%
  \institution{Shanghai Jiao Tong University}
  \city{Shanghai}
  \country{China}
}

\renewcommand{\shortauthors}{Xiong et al.}

\begin{abstract}
Embodied navigation (EN) advances traditional navigation by enabling robots to perform complex egocentric tasks through sensing, social, and motion intelligence. In contrast to classic methodologies that rely on explicit localization and pre-defined maps, EN leverages egocentric perception and human-like interaction strategies. This survey introduces a comprehensive EN formulation structured into five stages: Transition, Observation, Fusion, Reward-policy construction, and Action (TOFRA). The TOFRA framework serves to synthesize the current state of the art, provide a critical review of relevant platforms and evaluation metrics, and identify critical open research challenges. A list of studies is available at \url{https://github.com/Franky-X/Awesome-Embodied-Navigation}.
\end{abstract}




\maketitle

\section{Introduction}

Embodied artificial intelligence (EAI) was first proposed by Alan Turing in \cite{EAI_Turing} as a concept that involves learning through egocentric sensors and human interaction for intelligent robotic systems. In contrast to large artificial intelligence models for language or vision such as \cite{Attention, BERT, LoRA, GPT4}, EAI emphasizes knowledge acquisition from an egocentric perspective and interaction with humans or the surrounding environment. It is regarded as a promising approach for achieving artificial general intelligence (AGI) \cite{AGI_Scene}.

{Building upon EAI, the concept of embodiment extends naturally to robotic navigation systems. These systems inherently exhibit embodiment through their egocentric perception and distributed edge computing capabilities. The development of intelligence and adaptive skills in such navigation contexts fundamentally aligns with EAI paradigms. While traditional navigation research has primarily focused on state estimation, point-to-point path planning, and optimal control strategies, recent advances in artificial intelligence have elevated navigation to an intelligence-driven paradigm, giving rise to \textbf{Embodied Navigation (EN)}. EN encompasses the development of advanced navigation capabilities through the integration of sensing, social, and motion intelligence within autonomous agents. Compared to conventional approaches, such as Global Navigation Satellite System (GNSS) with pre-existing global maps and classical Simultaneous Localization and Mapping (SLAM) methodologies, EN demonstrates several distinctive characteristics:}

\begin{enumerate}
    \item \textbf{Sensing Intelligence:} EN systems achieve autonomous spatial understanding via multimodal \textit{egocentric active perception}, integrating proprioceptive and exteroceptive sensors instead of relying on pre-built global maps.
    \item \textbf{Social Intelligence:} EN systems interpret \textit{high-level human instructions} through language and semantic scene understanding to perform complex tasks, moving beyond predefined waypoints.
    \item \textbf{Motion Intelligence:} EN agents utilize \textit{adaptive and evolving motion skills} with high degrees of freedom (DoFs) for complex physical interaction, unlike traditional fixed-path following.
\end{enumerate}

\begin{figure}
    \centering
\subfloat[Process of EN]{\includegraphics[width=0.4\linewidth]{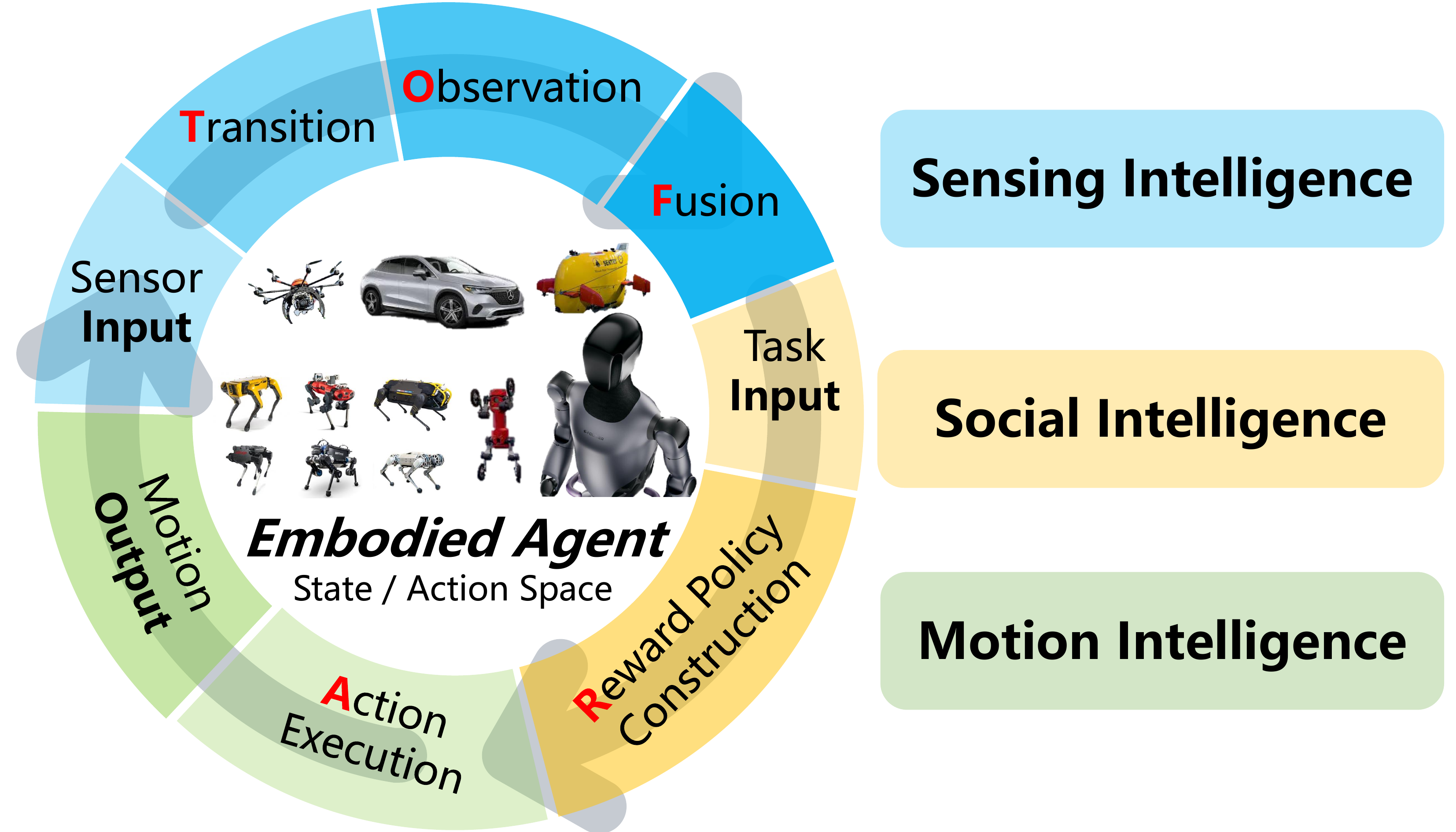}
    \label{fig:Structure}}
\subfloat[Paradigm of EN]{\includegraphics[width=0.55\linewidth]{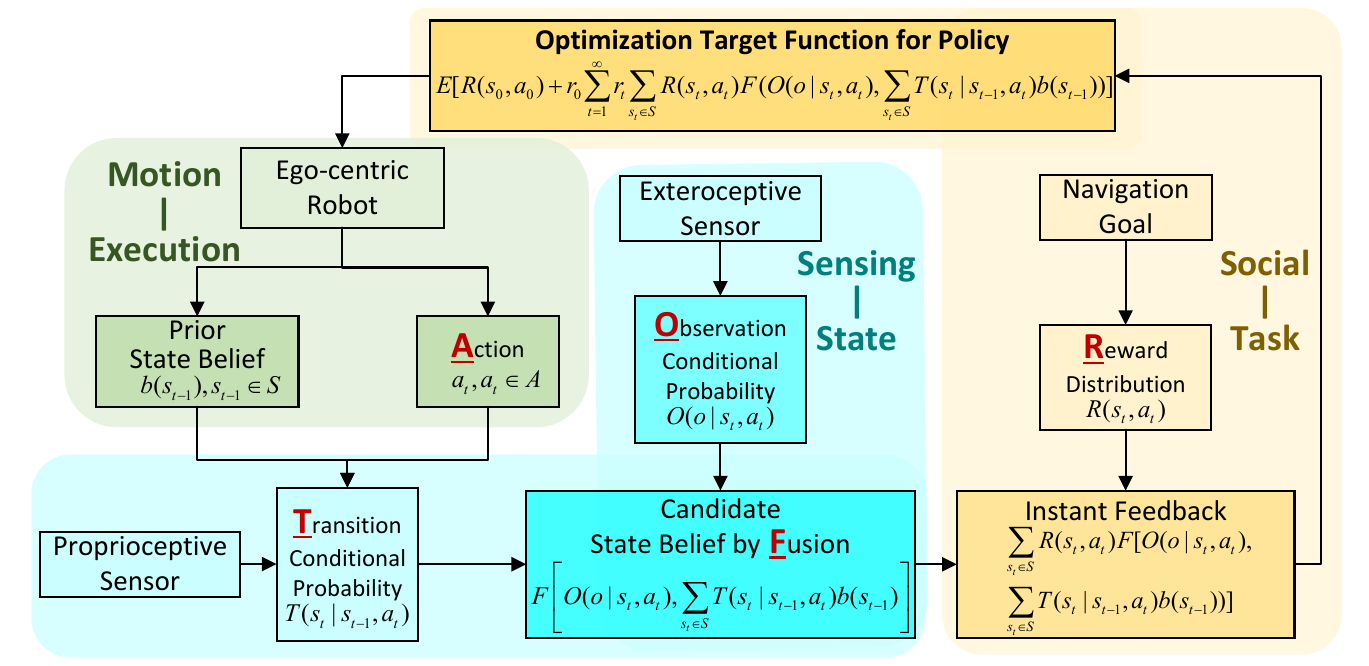}
    \label{fig:EN_Paradigm}}
    \vspace{-0.3cm}
    \caption{An EN system incorporates sensing, social, and motion intelligence. The EN process can be modeled through stages, including transition, observation, fusion of proprioception and exteroception, reward-policy construction, and action execution (TOFRA).}
    \vspace{-0.5cm}
\end{figure}

Recently, the EN problem has been addressed from different perspectives, primarily from computer vision (CV), classic robotics, and bionics. CV approaches \cite{CV_VXN, CV_EEN, CV_LMEN, CV_ESN} typically focus on \textit{Social Intelligence}, employing ideal agents with perfect state information and actions to solve complex language-based tasks. Conversely, robotics methods \cite{RS_ASLAM, RO_AA, RO_PP, RO_UN} prioritize \textit{Sensing Intelligence}, developing robust, multi-modal systems for real-world exploration and point-goal navigation, although with less sophisticated task understanding. A third stream investigates neuromorphic approaches for bio-inspired navigation \cite{NeuroNav_Survey}.

\begin{table}[!t]
  \centering
  \caption{Comparison between this survey and recent previous ones.}
  \renewcommand{\arraystretch}{1.3}
  \setlength{\tabcolsep}{0.05cm}
  \resizebox{\textwidth}{!}{
    \begin{tabular}{c|cc|c|cc|c|ccc|cccc|cc|ccc}
    \hline
    & & & \multicolumn{1}{c|}{\textbf{Topic}} & \multicolumn{2}{c|}{\textbf{Framework}} & \multicolumn{1}{c|}{\textbf{Eval.}} & \multicolumn{3}{c|}{\textbf{Sensing Intelligence}} & \multicolumn{4}{c|}{\textbf{Social Intelligence}} & \multicolumn{2}{c|}{\textbf{Motion Intelligence}} & \multicolumn{3}{c}{\textbf{Application Scenarios}} \\
    \cline{4-19}
    \textbf{Year} & \textbf{Ref.} & \textbf{Pub.} & \makecell{Main\\Focus} & \makecell{Framework\\Formulation} & \makecell{Unified\\Sum.} & \makecell{Metrics} & \makecell{\textbf{Proprio-}\\cept.} & \makecell{Extero-\\cept.} & \makecell{\textbf{Fusion}\\SLAM} & \makecell{Image\\Goal} & \makecell{Object\\Goal} & \makecell{Visual\\Lang.} & \makecell{Explor-\\ation} & \makecell{\textbf{Motion}\\Skills} & \makecell{\textbf{Morpho-}\\logical} & \makecell{Auton.\\Driving} & \makecell{Indoor\\Nav.} & \makecell{\textbf{Terrain}\\Nav.} \\
    \hline
    2019 & \cite{SV_IIIN} & \textbf{CSUR} & Indoor Nav. & $\times$ & $\times$ & $\times$ & $\times$ & \checkmark & $\times$ & $\times$ & $\times$ & $\times$ & $\times$ & $\times$ & $\times$ & $\times$ & \checkmark & $\times$ \\
    2020 & \cite{SV_AVN} & \textbf{CSUR} & Visual Nav. & $\times$ & $\times$ & $\times$ & $\times$ & \checkmark & \checkmark & $\times$ & $\times$ & $\times$ & \checkmark  & $\times$ & $\times$ & $\times$ & \checkmark & $\times$ \\
    2021 & \cite{RS_EVN} & ArXiv & VLN & $\times$ & $\times$ & \checkmark & $\times$ & \checkmark & $\times$ & \checkmark & \checkmark & \checkmark & $\times$ & $\times$ & $\times$ & \checkmark & \checkmark & $\times$ \\
    2022 & \cite{RS_EAI_FS2RT} & \textbf{TETCI} & EAI & $\times$ & $\times$ & \checkmark & $\times$ & \checkmark & $\times$ & \checkmark & \checkmark & \checkmark & \checkmark & $\times$ & $\times$ & $\times$ & \checkmark & $\times$ \\
    2023 & \cite{RS_VN_EAI} &\textbf{EAAI} & Visual Nav. & $\times$ & $\times$ & \checkmark & $\times$ & \checkmark & $\times$ & \checkmark & \checkmark & \checkmark & \checkmark & \checkmark & $\times$ & \checkmark & \checkmark & $\times$ \\
    2023 & \cite{RS_EN_LLM} & ArXiv & LLM for EN & $\times$ & $\times$ & \checkmark & $\times$ & \checkmark & $\times$ & \checkmark & \checkmark & \checkmark & \checkmark & $\times$ & $\times$ & \checkmark & \checkmark & $\times$ \\
    2023 & \cite{RS_ASLAM} & \textbf{TRO} & Active SLAM & \checkmark & $\times$ & $\times$ & $\times$ & \checkmark & \checkmark & $\times$ & $\times$ & $\times$ & \checkmark & $\times$ & $\times$ & $\times$ & \checkmark & $\times$ \\
    2024 & \cite{RS_EAI_VLA} & ArXiv & VLA & $\times$ & $\times$ & $\times$ & $\times$ & \checkmark & $\times$ & \checkmark & \checkmark & \checkmark & $\times$ & \checkmark & $\times$ & $\times$ & \checkmark & $\times$ \\
    2024 & \cite{RS_OGN} & \textbf{TASE} & Object Nav. & $\times$ & $\times$ & \checkmark & $\times$ & \checkmark & $\times$ & $\times$ & \checkmark & \checkmark & \checkmark & $\times$ & $\times$ & $\times$ & \checkmark & $\times$ \\
    2024 & \cite{NeuroNav_Survey} & \textbf{CSUR} & Neuro Nav. & $\times$ & $\times$ & $\times$ & $\times$ & \checkmark & \checkmark & $\times$ & $\times$ & $\times$ & \checkmark & $\times$ & $\times$ & $\times$ & \checkmark & $\times$ \\
    2025 & \cite{EN_China_survey} & \textbf{SCIS} & General EN & $\times$ & $\times$ & $\times$ & $\times$ & \checkmark & \checkmark & \checkmark & \checkmark & \checkmark & \checkmark & \checkmark & \checkmark & \checkmark & \checkmark & \checkmark \\
    2025 & \cite{RNM_Phy_survey} & ArXiv & Nav. and Mani. & $\times$ & $\times$ & \checkmark & $\times$ & \checkmark & $\times$ & \checkmark & \checkmark & \checkmark & $\times$ & \checkmark & $\times$ & \checkmark & \checkmark & $\times$ \\
    2025 & \cite{SV_AN} & \textbf{CSUR} & Auto. Nav. & $\times$ & $\times$ & \checkmark & $\times$ & \checkmark & \checkmark & \checkmark & \checkmark & \checkmark & $\times$ & \checkmark & $\times$ & \checkmark & \checkmark & $\times$ \\
    \hline
    - & \textbf{Ours} & - & \textbf{Unified EN} & \checkmark & \checkmark & \checkmark & \checkmark & \checkmark & \checkmark & \checkmark & \checkmark & \checkmark & \checkmark & \checkmark & \checkmark & \checkmark & \checkmark & \checkmark \\
    \hline
    \end{tabular}
  }
  \label{tab:survey_comprison}
\end{table}

Existing surveys on EN typically reflect the priorities of their originating fields. Surveys from the CV community mainly focus on \textit{Social Intelligence} (e.g., visual-language navigation) in idealized geometric settings, often abstracting away real-world sensing and motion challenges \cite{RS_EVN, RS_EAI_VLA, RS_EN_LLM}. Conversely, classic robotics surveys primarily address \textit{Sensing Intelligence} (e.g., active SLAM) while typically overlooking advanced social cognition and complex motion skills \cite{RS_ASLAM, SV_AN}. Additionally, neuromorphic surveys examine bio-inspired mechanisms but exist similar gaps in social and motion intelligence coverage \cite{NeuroNav_Survey}. While recent works have cataloged EN modules \cite{EN_China_survey, RNM_Phy_survey}, they lack a unifying structure to integrate these disparate research efforts. To address this limitation, our survey introduces the \textit{TOFRA} framework: a unified formulation that systematically integrates contributions from all fields into a coherent structure, as detailed in Table \ref{tab:survey_comprison}.

This survey presents a unified formulation of the EN problem and systematically integrates current research within this framework. First, we formulate the process of EN as a general probabilistic model using a partially observable Markov decision process (POMDP). We then divide the unified EN model into five stages: \underline{T}ransition, \underline{O}bservation, \underline{F}usion of proprioception and exteroception, \underline{R}eward-policy construction, and \underline{A}ction execution (TOFRA), which serves as a structure to integrate existing research. Second, we summarize and categorize existing EN-related methods according to these five stages. Third, we review platforms and evaluation metrics developed for EN tasks. Finally, we identify open research problems to highlight promising directions for future work.
Our contributions are summarized as follows:
\begin{enumerate}
    \item A comprehensive formulation of the EN problem, divided into five distinct stages for clear integration of current research.
    \item A unification and systematic categorization of existing methods from various research fields within the proposed TOFRA framework.
    \item An identification and discussion of open research problems to guide the future development of practical and general EN systems.
\end{enumerate}

The remainder of this survey is organized as follows: Section 2 presents the unified formulation of EN and divides the process into five distinct stages. Sections 3-8 summarize and integrate current works related to EN according to these five stages. Section 9 introduces systematic EN approaches applied across three representative navigation scenarios. Section 10 discusses existing simulation and real-world platforms for EN tasks. Section 11 identifies open research problems and future directions. Finally, Section 12 concludes the paper with a summary.



\begin{figure*}
    \centering
    \includegraphics[width=1.0\linewidth]{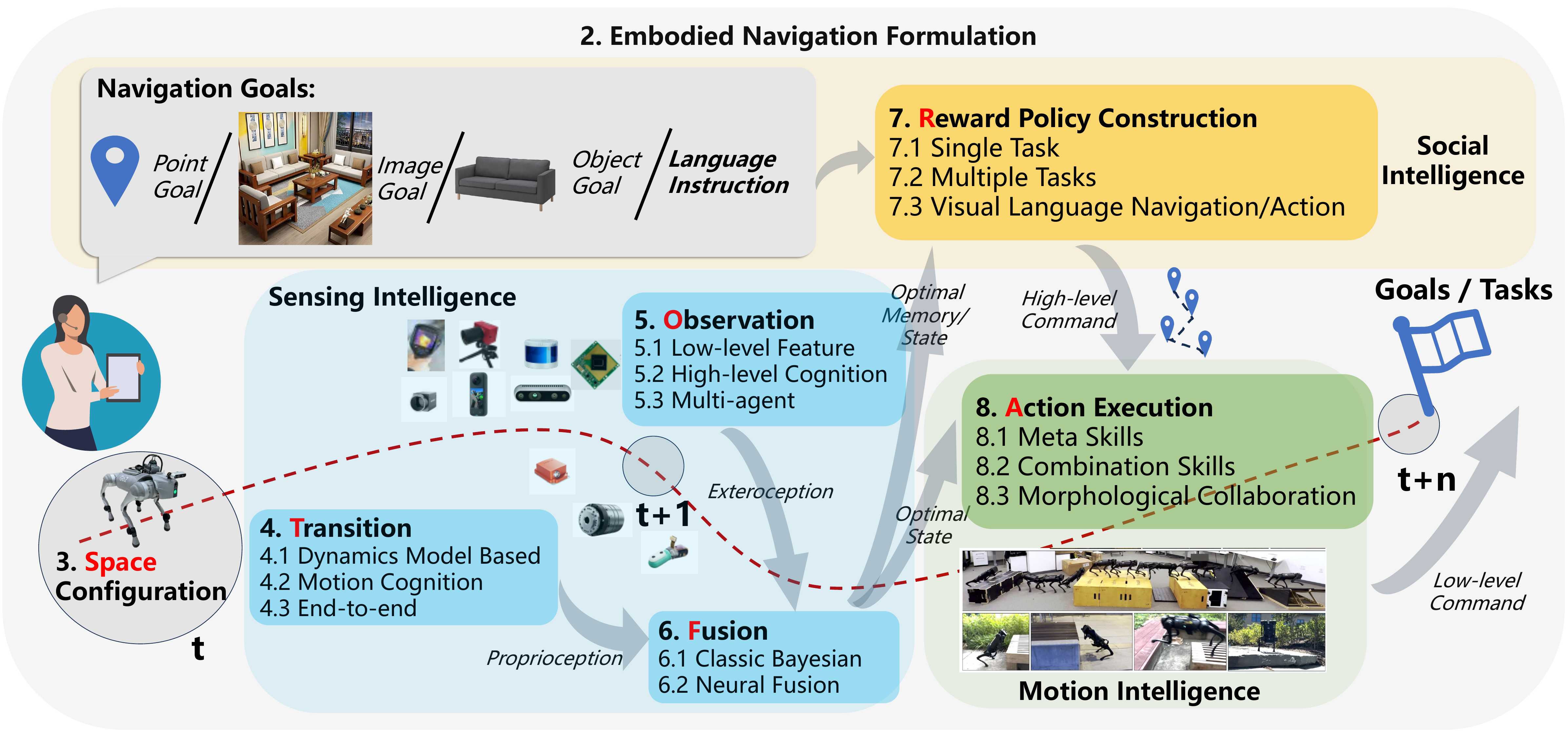}
    \caption{The pipeline and stages of a typical EN system. Current studies focusing on different stages of the EN system are summarized into the TOFRA framework in this survey.}
    \label{fig:ENFramework}
    \vspace{-0.5cm}
\end{figure*}

\section{EN Problem Formulation}
This section presents a unified formulation of the EN problem and systematically divides the navigation process into prerequisites and five sequential stages. This structured approach enables us to coherently integrate and analyze existing EN-related research within a comprehensive framework.
\subsection{Problem Formulation}
The aim of EN is to efficiently accomplish advanced navigation tasks in unknown environments using robotic systems equipped with egocentric sensors. The decision-making process in EN can be formulated within the general POMDP framework\cite{POMDP}. Fig. \ref{fig:ENFramework} illustrates the pipeline and stages of a typical EN system, as detailed in the following sections.

Firstly, an EN system must maintain a memory of state that includes both robotic and environmental states, enabling the robot to acquire knowledge about its position and surroundings. Due to uncertainty introduced by noise, the state is modeled as a probability distribution in a belief space, which is given by
\begin{equation}
\begin{aligned}
    b(\mathbf{s}_t) &: \mathbf{s}_t \in \mathcal{S}, \\
    \text{s.t} &\sum_{\mathbf{s}_t\in \mathcal{S}} b(\mathbf{s}_t) = 1,
\end{aligned}
\end{equation}
where $\mathbf{s}_t$ is the robot state at the time $t$, $b(\mathbf{s}_t)$ is the probability distribution of $\mathbf{s}_t$ and $\mathcal{S}$ is the space of all possible robot states.

Secondly, action is another critical element of the EN system, which allows a robot to move within the environment to approach its navigation goal. It is defined by \warning{$\mathbf{a}_t \in \mathcal{A}$,
where $\mathbf{a}_t$ is the action that the robot executes at time $t$, and $\mathcal{A}$ is the space of all possible actions available to the robot.}

By applying action $\mathbf{a}_t$, $\mathbf{s}_{t-1}$ moves to the next state $\mathbf{s}_t$. Given the probability distribution $b(\mathbf{s}_{t-1})$ at time $t-1$ and the action $\mathbf{a}_t$, one can deduce the probability distribution $\hat b(\mathbf{s}_{t})$ at time $t$ using a stochastic state transition model $T(\mathbf{s}_t|\mathbf{s}_{t-1},\mathbf{a}_t)$, which is defined by
\begin{equation}
   \hat{b}(\mathbf{s}_t) = \sum_{\mathbf{s}_{t-1} \in \mathcal{S}}T(\mathbf{s}_t|\mathbf{s}_{t-1},\mathbf{a}_t) b(\mathbf{s}_{t-1}), 
\end{equation}
where $\hat{b}(\mathbf{s}_t)$ is the prior probability distribution of state $\mathbf{s}_t$. The state transition model is typically derived from the action instructions and the proprioceptive sensors. This process is the formulation of \textit{proprioceptive intelligence} for an agent.

After obtaining the prior state distribution $\hat{b}(\mathbf{s}_t)$, the robot can acquire environmental observations through ego-perceptive sensors. These observations can be represented using a \warning{probabilistic observation model} $O(\mathbf{o}_t|\mathbf{s}_{t},\mathbf{a}_t)$, with which the marginal observation probability can be expressed as
\begin{equation}
    O(\mathbf{o}_t) = \sum_{\mathbf{s}_t\in \mathcal{S}} O(\mathbf{o}_t|\mathbf{s}_t,\mathbf{a}_t)\hat{b}(\mathbf{s}_t). 
\end{equation}
This process constitutes the formulation of \textit{exteroceptive intelligence} for an agent.

After obtaining state transition and environmental observation information, these two sources can be fused to determine the optimal probability distribution of state $\mathbf{s}_t$. This fusion process is formulated as
\begin{equation}
    b(\mathbf{s}_t) = F( O(\mathbf{o}_t|\mathbf{s}_t,\mathbf{a}_t), \hat{b}(\mathbf{s}_{t})),
\end{equation}
where $F$ denotes the fusion algorithm that combines the information from $\hat{b}(\mathbf{s}_{t})$ and $O(\mathbf{o}_t|\mathbf{s}_t,\mathbf{a}_t)$. This fusion algorithm may implement either a classical Bayesian approach or utilize learning-based implicit methods. This process represents the formulation of sensing fusion intelligence for an agent. Together, the transition, observation, and fusion processes constitute the complete \textit{sensing intelligence} framework.

Once the action is executed and the optimal state distribution is computed, the contribution of the action and current state to the navigation goal is assessed. This assessment is conducted using a reward distribution function $R(\mathbf{s}_t,\mathbf{a}_t)$, which quantifies the benefit of achieving the navigation task. The total expected reward $r(b,\mathbf{a}_t)$ associated with the current state distribution and action can be computed as
\begin{equation}
    r(b,\mathbf{a}_t) = \sum_{\mathbf{s}_t \in \mathcal{S}} R(\mathbf{s}_t,\mathbf{a}_t)b(\mathbf{s}_t).
\end{equation}

\warning{Then a policy function $\pi(\mathbf{s}_t)$ is formulated to generate
the desired action sequence.} The objective of an EN system is to maximize the expected accumulated reward \warning{$r(b,\mathbf{a}_t)$} \warning{over the state probability distribution $b$}; that is 
\begin{equation}
    \warning{ \pi^* = \mathop{\arg\max}\limits_{\pi} \mathbb{E}\left[\sum_{t=0}^\infty \gamma^t \cdot r(b,\mathbf{a}_t)\right], }
\label{eq:TF}
\end{equation}
where $\gamma^t$ is the discount factor that balances immediate and future rewards. The optimal policy $\pi^*$ generates the action sequence $\{\mathbf{a}_0,\mathbf{a}_1,\mathbf{a}_2,...\}$ that maximizes the expected cumulative reward. This process is the formulation of \textit{social intelligence} for an agent.
\warning{Finally, the agent follows the action sequence with the motion skills. This process is the formulation of \textit{motion intelligence} for an agent.} 


\subsection{Five Stages of EN Problem}
To clearly divide the entire EN problem into subtasks for resolution, the target function Eq. (\ref{eq:TF}) is further elaborated as
\begin{equation}
\begin{aligned}
    \mathbb{E}\left[\sum_{t=0}^\infty \gamma^t \cdot r(b,\mathbf{a}_t)\right] = \mathbb{E}\Bigg[\sum_{\mathbf{s}_0 \in \mathcal{S}} R(\mathbf{s}_0,\mathbf{a}_0)b(\mathbf{s}_0) + \\
    \sum_{t=1}^\infty \gamma^t \cdot \sum_{\mathbf{s}_t \in \mathcal{S}} R(\mathbf{s}_t,\mathbf{a}_t)F\Big(O(\mathbf{o}_t|\mathbf{s}_t,\mathbf{a}_t), \sum_{\mathbf{s}_{t-1} \in \mathcal{S}}T(\mathbf{s}_t|\mathbf{s}_{t-1}, \mathbf{a}_t)b(\mathbf{s}_{t-1})\Big)\Bigg].
\end{aligned}
\label{eq:TF_E}
\end{equation}
From the elaborated target function Eq. (\ref{eq:TF_E}), the elements and stages of an EN system can be understood, as shown in Fig. \ref{fig:ENFramework}.
Firstly, the EN system requires the configuration of state modeling and action capabilities to define the state and action spaces, as well as the initial state distribution $b(\mathbf{s}_0)$. Then, the EN system operates through the following stages:
\begin{enumerate}
        \item Transition (T). The key aspect of this stage is to configure $T(\mathbf{s}_t|\mathbf{s}_{t-1},\mathbf{a}_t)$, so that the prior $\hat{b}(\mathbf{s}_{t})=\sum_{\mathbf{s}_{t-1} \in \mathcal{S}} T(\mathbf{s}_t|\mathbf{s}_{t-1},\mathbf{a}_t) b(\mathbf{s}_{t-1})$ is computed.
        \item Observation (O). In this stage, the EN system acquires environmental observation through ego-perceptive sensors to determine $O(\mathbf{o}_t|\mathbf{s}_t,\mathbf{a}_t)$.
        \item Fusion of proprioception and exteroception (F). This stage fuses the information from $\hat{b}(\mathbf{s}_{t})$ and $O(\mathbf{o}_t|\mathbf{s}_t,\mathbf{a}_t)$ to compute the optimal state distribution $b(\mathbf{s}_t) = F(O(\mathbf{o}_t|\mathbf{s}_t,\mathbf{a}_t), \hat{b}(\mathbf{s}_{t}))$.
        \item Task reward-policy construction (R). The EN system evaluates the contribution of the current state and action to the navigation goal by calculating the reward $r(b,\mathbf{a}_t) = \sum_{\mathbf{s}_t \in \mathcal{S}} R(\mathbf{s}_t,\mathbf{a}_t)b(\mathbf{s}_t)$. Then, a policy model $\pi(\mathbf{s}_t)$ is formulated to generate an action sequence that maximizes the objective $ \mathbb{E}[\sum_{t=0}^\infty \gamma^t \cdot r(b,\mathbf{a}_t)]$.
        \item Action execution (A). Finally, the EN system executes the action sequence $\{\mathbf{a}_0,\mathbf{a}_1,\mathbf{a}_2,...\}$ generated by the policy, utilizing the agent's motion skills to efficiently achieve the navigation goals.
\end{enumerate}
As illustrated in Fig. \ref{fig:EN_Paradigm}, the TOFRA framework provides a structured approach to understanding the complete EN process. In Sections 3-9, we systematically review and integrate current research across each stage of this framework, demonstrating how the field has evolved along the dimensions of sensing, social, and motion intelligence.

\section{State and Action Space Configuration: EN Prerequisites}
In this section, the configurations of state and action spaces are introduced, which determine the computational framework and the solution space of the EN problem.

\begin{table*}
\centering
\caption{EN Space Configurations and Their Advantages, Limitations and Applications}
\renewcommand{\arraystretch}{1.5} 
\setlength{\tabcolsep}{0.1cm} 
\resizebox{0.8 \textwidth}{!}{
\begin{tabular}{c|c|c|c|c|c|c}
\hline
\textbf{\makecell{EN \\ Space}} & \textbf{Modules} & \textbf{Categories} & \textbf{Advantages} & \textbf{Limitations} & \textbf{Applications} & \textbf{References} \\ \hline

\multirow{8}{*}{\makecell{State \\ Space}} 
& \multirow{4}{*}{Body} 
& Vector & Easy to Extend & Inconsistency Issues & General State Estimation & \cite{SP_Vector_OpenVINS, SP_Vector_VINS_Mono} \\ \cline{3-7}
& & Lie Group & High Consistency & Parameter Sensitivity & Filter-Based Estimation & \cite{SP_InEKF_OB, SP_PIEKF, SP_InEKF} \\ \cline{3-7}
& & Quaternion-Based & Consistent and Efficient & Parameter Sensitivity & Filter-Based Estimation & \cite{SP_TriQ} \\ \cline{3-7}
& & Implicit & High Representation Capability & Poor Generalizability & Learning-Based Navigation & \cite{SP_VXN, SP_MTR, SP_SVN} \\ \cline{2-7}

& \multirow{4}{*}{Environment} 
& Point Cloud & Dense and Direct Mapping & Incomplete Coverage & SFM, Classic Odometry & \cite{SP_ENV_PCL} \\ \cline{3-7}
& & Voxel & Semi-Dense and Direct & Loss of Fine Details & Occupancy Estimation & \cite{SP_ENV_OCC} \\ \cline{3-7}
& & Implicit & High Representation Capability & Poor Generalizability & Learning-Based Navigation & \cite{SP_ENV_IMP} \\ \cline{3-7}
& & Graph & Sparse and Efficient & Limited Geometric Detail & Bio-inspired Navigation & \cite{BIG} \\ \hline

\multirow{6}{*}{\makecell{Action \\ Space}} 
& \multirow{4}{*}{\makecell{Low-level \\ Control}} 
& Ground Wheel & Easy to Deploy & Poor Traversability & Flat Ground Navigation & \cite{AC_KITTI, AC_CARLA} \\ \cline{3-7}
& & UAVs/AUVs & Flexible Mobility & Limited Payload Capacity & Aerial/Marine Navigation & \cite{AC_UAV, AC_AUV} \\ \cline{3-7}
& & Legged Robots & Robust Terrain Handling & Complex Control & Rough Terrain Navigation & \cite{AC_ANY, AC_MIT, AC_ANY_3D, AC_ALOC, AC_Humanoid} \\ \cline{3-7}
& & Wheel-Legged & Efficient and Flexible & Limited Stability & Semi-Terrain Navigation & \cite{AC_WL, AC_WL_SR} \\ \cline{2-7}

& \multirow{2}{*}{\makecell{High-level \\ Command}} 
& Legged Robots & Simple Interface & Limited Flexibility & Terrain Navigation & \cite{AC_LR_HIGH} \\ \cline{3-7}
& & Wheel-Legged & Simple Interface & Limited Flexibility & Semi-Terrain Navigation & \cite{AC_WL_SR} \\ \hline

\end{tabular}
}
\label{table:SPACE}
\end{table*}

\subsection{State Space}
The state space defines the possible configurations of both the robot and its environment. It serves as the foundation for computational frameworks of transition, observation, and fusion algorithms.
The representation of state is fundamental to the effectiveness of EN systems\cite{SP_Basic}.

\subsubsection{Explicit Representation}
Explicit representation methods are employed to describe an agent's state, including three-dimensional (3-D) position, rotation, and environmental features. These methods characterize the state using variables with precise physical interpretations. Several typical approaches to explicit representation are described below.

\textit{Concatenate Vector:} A common method is to represent the state as a concatenated vector \cite{SP_Vector_OpenVINS, SP_Vector_VINS_Mono}:
\begin{equation}
    \chi = [\mathbf{p}, \mathbf{v}, \mathbf{q}, \mathbf{f}_1, ..., \mathbf{f}_n, \mathbf{t}_1, ..., \mathbf{t}_n],
\end{equation}
where $\mathbf{p}, \mathbf{v}, \mathbf{q}$ are the robot's position, velocity, and orientation; $\mathbf{f}_i$ are environmental features \cite{SP_ENV_PCL, SP_ENV_OCC}; and $\mathbf{t}_i$ are other system parameters. \warning{However, this method can lead to estimation inconsistencies because sub-state errors are computed in non-uniform coordinate systems \cite{SP_InEKF}.}

\textit{Lie Group:} To improve estimation consistency, the state can be represented on a Lie group \cite{SP_InEKF_OB}. This approach leverages the group-affine property to make Jacobian computations independent of the current state estimate, mitigating linearization errors in optimization \cite{SP_InEKF_OB, SP_PIEKF}. The state $\chi_I$, including rotation $\mathbf{R}$, velocity $\mathbf{v}$, position $\mathbf{p}$, and other parameters $\mathbf{t}_i$, can be structured as:
\begin{equation}
\chi_I=\left[\begin{array}{ccccccc}
\mathbf{R} & \mathbf{v} & \mathbf{p} & & & & \\
& 1 & & & & & \\
& & 1 & & & & \\
& & & \mathbf{I}_n & \mathbf{t}_1 & & \\
& & & & \ddots & \ddots & \\
& & & & & \mathbf{I}_n & \mathbf{t}_n \\
& & & & & & 1
\end{array}\right].
\end{equation}
This formulation is particularly effective in inertial navigation systems for enhancing stability and consistency \cite{SP_Lie_Huang0, SP_Lie_Huang1, SP_Lie_Huang2}.


\textit{Quaternion-Based Representation:} Quaternions serve as a fundamental method for representing rotation in three-dimensional space. Dual quaternions extend this framework to incorporate translation alongside rotation\cite{SP_DQ}. A recent advancement in this domain is the trident quaternion \cite{SP_TriQ}, which further enhances the quaternion concept by integrating translation, velocity, and rotation into a unified representation. The mathematical formulation of a trident quaternion is given by
\begin{equation}
\breve{q}=q+\varepsilon_1 q^{\prime}+\varepsilon_2 q^{\prime \prime},
\end{equation}
where $\breve{q}$ represents the complete robot state. The components $q, q^{\prime}, q^{\prime \prime}$ are traditional quaternions\cite{SP_Basic} representing. The dual numbers $\varepsilon_1$ and $\varepsilon_2$ satisfy $\varepsilon_1^2=\varepsilon_2^2=\varepsilon_1 \varepsilon_2=0$, with $\varepsilon_1 \neq 0$ and $\varepsilon_2 \neq 0$. \warning{Trident quaternions unify the coordinate system for error state computation, making the Jacobians independent during the optimization process. This property is similar to Lie group representations and ensures consistency during state estimation.}

\subsubsection{Implicit Representation}
In contrast to the explicit representation methods discussed above, implicit representation characterizes the state using abstract, uninterpretable feature vectors. 
This technique is widely employed in deep learning-based navigation frameworks, where the state is represented as a learned feature vector; that is
\begin{equation}
\mathbf{s}_t = f_\theta(\mathbf{o}_t),
\end{equation}
where $\mathbf{s}_t$ is the defined state at time $t$. $f_\theta(\cdot)$ is the embedding function parameterized by $\theta$, which contains the weights of the neural network, and $\mathbf{o}_t$ is the observation of the robot at time $t$. This state representation approach is learned implicitly during the training process of neural networks within specific scenarios\cite{SP_VXN, SP_MTR, SP_SVN, SP_ENV_IMP}.

Another branch of implicit representation research focuses on bio-inspired cell-based methods. In \cite{NeuroSLAM}, the state space is configured using bio-inspired cells to represent both agent and environmental states. This form of representation is computationally efficient and particularly suitable for implementation on neuromorphic devices \cite{ResonatorOdo}.

\subsubsection{Discussion and Trends}

State space representation has evolved through several distinct developmental stages:

\begin{itemize}
    \item \textbf{From \textbf{Vector Representation} to \textbf{Geometric Consistency}:} Evolution from simple concatenated vectors toward \textbf{Lie groups} and \textbf{quaternion formulations} that maintain geometric integrity through principled transformations.
    
    \item \textbf{From \textbf{Explicit} to \textbf{Implicit Representations}:} Transition from \textbf{interpretable mathematical models} to flexible \textbf{learning-based representations} that better adapt to unstructured environments.
    
    \item \textbf{Toward \textbf{Neuromorphic Efficiency}:} Emergence of \textbf{brain-inspired approaches} offering computationally efficient representations that balance \textbf{mathematical rigor} with \textbf{adaptive learning} capabilities.
\end{itemize}





\subsection{Action Space}
The action space is determined by an agent's physical capabilities, such as its locomotion system and degrees of freedom. As robotic control has advanced, action spaces have evolved to handle more diverse and complex environments. This subsection reviews the action spaces for prevalent embodied agent platforms.

\subsubsection{Ground Wheeled Vehicles} Ground wheeled vehicles represent the fundamental form of robotic agents for embodied navigation. The majority of EN research has focused on these platforms, which typically possess 3 DoFs\cite{AC_KITTI, AC_CARLA}. 
For implementation purposes, their continuous action space can often be simplified into discrete commands, such as moving forward, backward, turning left, and turning right, making them accessible platforms for algorithm development.

\subsubsection{Unmanned aerial vehicles (UAVs) and autonomous underwater vehicles (AUVs)} In contrast to ground wheeled vehicles, UAVs and AUVs operate with enhanced mobility, possessing 6 DoFs for navigation tasks \cite{AC_UAV, AC_AUV}. These degrees of freedom comprise three translational movements and three rotational movements, enabling full spatial maneuverability in aerial or aquatic environments.

\subsubsection{Legged Robots} Legged robots have been developed to navigate complex terrains with greater flexibility and versatility than conventional wheeled systems \cite{AC_ANY, AC_MIT}. The action space of legged robots can either be simplified to match ground wheeled vehicles \cite{AC_ANY_3D}, or defined in the original joint space with significantly higher degrees of freedom. For instance, quadrupedal robots typically operate with over 12 DoFs \cite{AC_ALOC}, while humanoid robots possess more than 25 DoFs \cite{AC_Humanoid} for locomotion during navigation tasks.

\subsubsection{Wheel-Legged Robots} Wheel-legged robots have gained popularity by combining the advantages of wheeled vehicles and legged systems, with their locomotion extensively studied \cite{AC_WL,AC_WL_SR}. This hybrid design distinguishes itself from traditional platforms by enabling both rapid traversal on moderate surfaces and agile movement across challenging terrains. The action space of these robots is typically represented as a 16-dimensional vector comprising commands for 12 joint positions (leg components) and 4 wheel velocities (wheel components)\cite{AC_WL_SR}.



    
    


\begin{figure}
    \centering
\subfloat[State transition pipeline and trend.]{\includegraphics[width=0.4\linewidth]{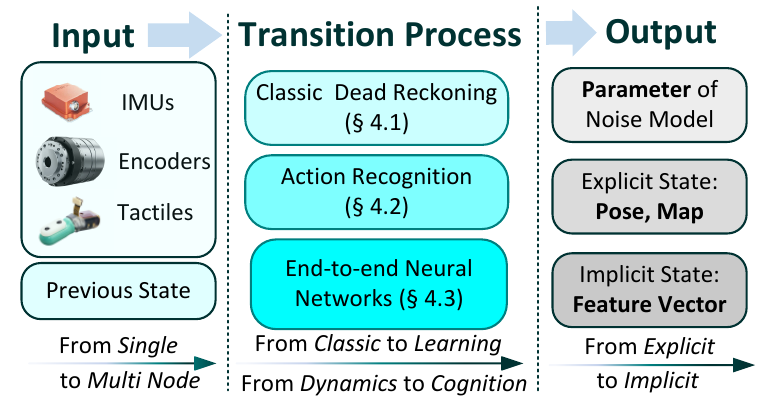}
    \label{fig:Structure}}
\subfloat[Observation pipeline and trend.]{\includegraphics[width=0.6\linewidth]{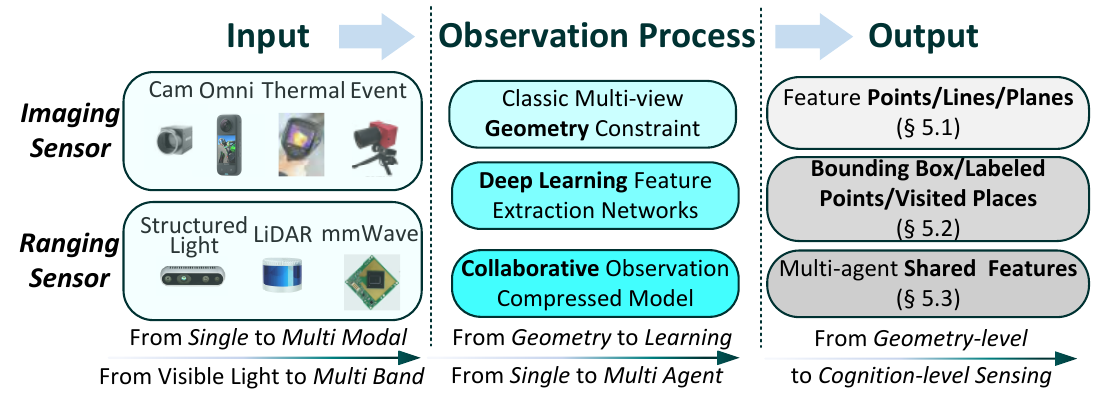}
    \label{fig:Structure}}
    \vspace{-0.3cm}
    \caption{A summary of the state transition and observation pipeline.}
    \vspace{-0.3cm}
\end{figure}

\section{Transition: Proprioceptive Intelligence}
The transition stage forms an agent's proprioceptive intelligence by predicting the next state after an action using internal sensors like IMUs. This section reviews three levels of transition methods: dynamics-based models, motion cognition-enhanced approaches, and end-to-end learning techniques.

\begin{table*}
\centering
\caption{State Transition Methods and Their Properties, Challenges and Applications}
\renewcommand{\arraystretch}{1.5} 
\setlength{\tabcolsep}{0.1cm} 
\resizebox{0.8 \textwidth}{!}{
\begin{tabular}{c|c|c|c|c|c|c}
\hline
\textbf{\makecell{Transition \\ Levels}} & \textbf{Modules} & \textbf{Categories} & \textbf{Properties} & \textbf{Challenges} & \textbf{Applications} & \textbf{References} \\ \hline

\multirow{9}{*}{Dynamics} 
& \multirow{3}{*}{Pre-integration} 
& Trapezoidal & Efficient and Robust & Loss of Accuracy & Real-time Embedded Systems & \cite{SP_Vector_VINS_Mono} \\ \cline{3-7}
& & Runge-Kutta & Improved Accuracy & Noise Sensitivity & Navigation Systems & \cite{SP_Vector_OpenVINS} \\ \cline{3-7}
& & Polynomials & High Accuracy & Computationally Expensive & Spacecraft, Marine Systems & \cite{ST_D_iNav} \\ \cline{2-7}

& \multirow{4}{*}{Dynamics} 
& Wheel Vehicles & Simple 2-D & Slip Model & Flat Ground Navigation & \cite{ST_D_GV} \\ \cline{3-7}
& & UAV/AUVs & 3-D Space & External Disturbance & Aerial and Marine Systems & \cite{ST_D_UAV, ST_D_AUV} \\ \cline{3-7}
& & Legged Robots & High DoFs & Complex Dynamics Model & Rough Terrain Navigation & \cite{ST_D_Legged} \\ \cline{3-7}
& & Wheel-Legged & Hybrid Models & Stability Challenges & Semi-Rough Terrain & \cite{ST_D_WL} \\ \cline{2-7}

& \multirow{3}{*}{Multi-node} 
& Pedestrian & Non-Rigid & Parameter Sensitivity & Motion Tracking & \cite{ST_D_LLK, ST_D_SENSORNET} \\ \cline{3-7}
& & Ground Vehicles & Rigid & Slip Model & Autonomous Ground Vehicles & \cite{ST_D_MIGV} \\ \cline{3-7}
& & Legged Robots & Non-Rigid & Hard to Calibrate & Terrain Navigation & \cite{ST_D_MIPO} \\ \cline{1-7}

\multirow{4}{*}{\makecell{Motion \\ Cognition}} 
& \multirow{2}{*}{\makecell{Stationary \\ Detection}} 
& Classic Model & Simple & Limited Accuracy & Stable Motion Tracking & \cite{ST_MC_PDR_ZUPT_BACK, ST_MC_PDR_ZUPT} \\ \cline{3-7}
& & Deep Learning & High Accuracy & Training Requirement & Fierce Motion Tracking & \cite{ST_MC_AIIMU} \\ \cline{2-7}

& \multirow{2}{*}{\makecell{Activity \\ Recognition}} 
& IMUs Only & Lightweight & Limited Context Awareness & PDR Navigation  & \cite{ST_MC_MARS, ST_MC_ML} \\ \cline{3-7}
& & Multi Modalities & Improved Robustness & Complex Data Fusion & AR/VR Estimation  & \cite{ST_MC_ERGB, ST_MC_ERGB_BENCH, ST_MC_EVENT, ST_MC_ERGB_IMU2} \\ \cline{1-7}

\multirow{2}{*}{End-to-end} 
& \multirow{2}{*}{\makecell{Neural \\ Networks}} 
& IMU Denoising & Noise Reduction & Limited to Certain Sensors & Real-time systems & \cite{ST_E2E_Deep_Denoise, ST_E2E_DeepIMU, ST_E2E_DUET, ST_E2E_DeepCali} \\ \cline{3-7}
& & Direct End-to-end & High Performance & Training Requirement & Learning-based Systems & \cite{ST_E2E_DMN, ST_E2E_IMU_Poser, ST_E2E_DF} \\ \cline{1-7}

\end{tabular}
}
\label{table:ST}
\end{table*}

\subsection{State Transition by Dynamics Model: Classic Dead Reckoning}
Classic state transition methods fall within the domain of dead reckoning, an approach grounded in system dynamics that utilizes proprioceptive sensors such as IMUs to obtain parameters like linear acceleration and angular velocity.

\subsubsection{IMU Integration}
The fundamental step in state transition via dynamics involves the pre-integration of IMU measurements. To handle discrete IMU data samples, various methods have been proposed to approximate the continuous integrals. The trapezoidal rule, adopted in \cite{SP_Vector_VINS_Mono}, offers computational efficiency but limited precision. For higher accuracy integration, Runge-Kutta methods are implemented in \cite{SP_Vector_OpenVINS}. Further advances include iterative integration and Chebyshev polynomials, which maximize integration accuracy as demonstrated in \cite{ST_D_iNav}.

\subsubsection{Agent Dynamics}
State transition can be computed by integrating acceleration and angular velocity data from IMUs and joint encoders mounted on the agent's body. The dynamics of various agent types have been extensively studied for navigation and control purposes, including ground wheeled vehicles \cite{ST_D_GV}, UAVs \cite{ST_D_UAV}, AUVs \cite{ST_D_AUV}, legged robots \cite{ST_D_Legged}, and wheel-legged robots \cite{ST_D_WL}. These well-established dynamic models serve as the foundation for classic state transition approaches, effectively utilizing sensor-acquired parameters to predict the agent's next state within the navigation process.

\subsubsection{Multi-node Measurements}
In addition to a single IMU, state estimation through multi-node IMUs has also been investigated to enhance the precision of state transition, particularly in high-degree-of-freedom agents. The strategy of employing multi-node IMUs is prevalent in the field of pedestrian localization, as it utilizes pedestrian dynamics to construct constraints among IMU nodes \cite{ST_D_LLK,ST_D_SENSORNET}. This approach has also been integrated into the state transition models of ground vehicles \cite{ST_D_MIGV} and legged robots \cite{ST_D_MIPO}. Through the integration of multi-node IMU measurements and system dynamics, significantly improved state transition accuracy can be achieved.

\subsection{State Transition with Motion Cognition Enhancement: Activity Recognition}
Advancements in deep learning and pattern recognition have enabled the integration of motion cognition into state transition modeling. This enhancement, extensively studied within the field of activity recognition, incorporates high-level knowledge about an agent's motion status—such as zero-velocity states, humanoid activities, or legged robot gaits. The state transition matrix $T(\mathbf{s}_t|\mathbf{s}_{t-1},\mathbf{a}_t)$ can be further elaborated by
\begin{equation}
    T(\mathbf{s}_t|\mathbf{s}_{t-1},\mathbf{a}_t) = \sum_{j=1}^n T(\mathbf{s}_t|\mathbf{s}_{t-1},\mathbf{a}_t,m_j)p(m_j),
\end{equation}
where $p(m_j)$ denotes the probability of a specific motion cognition $m_j$, and $T(\mathbf{s}_t|\mathbf{s}_{t-1},\mathbf{a}_t,m_j)$ represents the state transition matrix conditioned on the motion cognition $m_j$.

\subsubsection{Basic Velocity Status Detection}
The most fundamental application of motion cognition-enhanced state transition is the Zero Velocity Update (ZUPT) technique \cite{ST_MC_ZUPT}. ZUPT enables an agent to recognize when it is stationary, allowing it to correct drift in the estimated state. Classic model-based methods have been widely employed for zero-velocity detection in pedestrian localization systems \cite{ST_MC_PDR_ZUPT_BACK, ST_MC_PDR_ZUPT}. Similarly, in autonomous robotic platforms, stationary state detection has been utilized to establish motion constraints for vehicles, as demonstrated in \cite{ST_MC_AIIMU, ST_ZUPT_Slip}.

\subsubsection{Agent Activity Recognition}
Deep learning methods have been widely applied to activity recognition, enhancing the transition process. For high-DoF agents such as legged robots, recognizing current motion patterns or gaits significantly improves state transition accuracy. While robot-specific activity recognition remains an emerging field, methods can be effectively adapted from the extensive research on Human Activity Recognition (HAR). Some approaches \cite{ST_MC_MARS, ST_MC_ML} utilize IMUs exclusively for activity recognition, while others \cite{ST_MC_ERGB, ST_MC_ERGB_BENCH, ST_MC_EVENT, ST_MC_ERGB_IMU2} incorporate egocentric cameras to enhance recognition performance. By applying these techniques to identify its own motion state, an EN agent can employ more precise, context-specific transition models, thereby improving odometry accuracy and overall state estimation.



\subsection{End-to-end State Transition: Implicit Dynamics Modeling}
While classic methods rely on precise parameters and accurate dynamic models, these parameters and models may deteriorate due to various environmental factors and sensor uncertainties. To address these limitations, end-to-end deep learning methods have been developed to implicitly model complex dynamics and mitigate errors that are difficult to capture with classical approaches.
\subsubsection{Neural Denoising}
Deep learning networks are employed to denoise IMU measurements by calibrating sensor biases in real-time. The CNN-based approach for IMU denoising is first proposed in \cite{ST_E2E_Deep_Denoise}, with subsequent advancements incorporating dilated CNNs \cite{ST_E2E_DUET}. Recognizing the temporal characteristics of IMU data, researchers have implemented specialized architectures such as long short-term memory (LSTM) and temporal convolutional networks (TCN) to further improve denoising performance \cite{ST_E2E_DeepIMU, ST_E2E_DeepCali}.
\subsubsection{Neural Transition}
Beyond denoising approaches, neural networks can completely replace traditional dynamics models to address their inherent inaccuracies. In these end-to-end architectures, data from multi-node IMUs is processed through neural network layers that implicitly learn the transition function without requiring explicit physical modeling. This approach enables direct state transition computation, overcoming the limitations of classical integration methods \cite{ST_E2E_DMN, ST_E2E_IMU_Poser, ST_E2E_DF}.

\subsection{Discussion and Trends}
The evolution of state transition approaches reveals three progression paths:

\begin{itemize}
    \item \textbf{From \textbf{Simple Node} to \textbf{Higher DoFs}:} Beginning with \textbf{single-point motion models} and evolving toward \textbf{high-dimensional, multi-DoF transition frameworks} capable of representing increasingly sophisticated agent morphologies.
    
    \item \textbf{From \textbf{Quantitative Transition} to \textbf{Cognitive Generation}:} Moving from \textbf{strongly metric, equation-driven} transition models toward \textbf{weakly metric, context-aware} frameworks that incorporate higher-level motion understanding and prediction.
    
    \item \textbf{From \textbf{Geometric Discrete} to \textbf{Physical Continuous}:} Transitioning from \textbf{geometric, discrete state updates} to \textbf{physically-grounded continuous transition models} that better represent natural agent movement and environmental interactions.
\end{itemize}

    
    


\section{Observation: Exteroceptive Intelligence}
The observation stage forms an agent's exteroceptive intelligence by perceiving the environment through external sensors. This section reviews observation methods across three levels: low-level feature extraction, high-level cognitive understanding, and multi-agent collaborative sensing.

\begin{table*}
\centering
\caption{Observation Methods and Their Properties and Challenges}
\renewcommand{\arraystretch}{1.3} 
\setlength{\tabcolsep}{0.15cm} 
\resizebox{0.8\textwidth}{!}{ 
\begin{tabular}{c|c|c|c|c|c}
\hline
\textbf{Level} & \textbf{Categories} & \textbf{Dimension} & \textbf{Properties} & \textbf{Challenges} & \textbf{References} \\ \hline

\multirow{4}{*}{Low-level Feat.} 
& \multirow{2}{*}{Point Feat.} 
& 2D & Lightweight & Textureless Scenes & \cite{OB_P_SIFT, OB_P_SuperPoint, OB_P_SuperGlue, OB_P_Loftr} \\ \cline{3-6}
& & 3D & Depth Consistent & Textureless Surfaces & \cite{OB_P_LOAM} \\ \cline{2-6}

& \multirow{2}{*}{Structural Feat.} 
& 2D & Geometric Stability & Cluttered Environments & \cite{OB_StructVIO, OB_L_GlueStick} \\ \cline{3-6}
& & 3D & Geometric Stability & Registration Errors & \cite{OB_P_LOAM, OB_L_3D} \\ \hline

\multirow{9}{*}{High-level Cognition} 
& \multirow{3}{*}{Object Detection} 
& 2D & Category-level Boxing & Occlusion, Scale Var. & \cite{OB_OD_YOLOv5, OB_OD_GCN, OB_OD_Weather} \\ \cline{3-6}
& & 3D & Spatial-aware Detection & Sparse Point Clouds & \cite{OB_OD_DCGNN, OB_OD_ASIF_Net} \\ \cline{3-6}
& & Fusion & Cross-modal Enhancement & Sparse Association & \cite{OB_OD_DCF, OB_OD_EPNet} \\ \cline{2-6}

& \multirow{3}{*}{Semantic Seg.} 
& 2D & Dense Pixel-level Labels & Limited Depth Context & \cite{OB_SS_FCN, OB_SS_DeepLab, OB_SS_SegFormer} \\ \cline{3-6}
& & 3D & Spatial Semantics & High Memory Usage & \cite{OB_SS_Pointnet2, OB_SS_KPConv, OB_SS_RandLA, OB_SS_PointTran, OB_SS_Pointbert} \\ \cline{3-6}
& & Fusion & Textual/Spatial Semantics & Sensor Calibration & \cite{OB_SS_MFFNet} \\ \cline{2-6}

& \multirow{3}{*}{Place Recognition} 
& 2D & Place Texture Aware & Viewpoint/Illumination Var. & \cite{OB_PR_Netvlad, OB_PR_R2former, OB_PR_Anyloc} \\ \cline{3-6}
& & 3D & View-invariant & Computationally Heavy & \cite{OB_PR_SeqOT, OB_PR_BEVPlace} \\ \cline{3-6}
& & Fusion & Rich Texture \& Structure & Data Association & \cite{OB_PR_AdaFusion, OB_PR_EI} \\ \hline

\multirow{6}{*}{Multi-agent} 
& \multirow{2}{*}{Ground-Ground} 
& 2D–2D & Multi-view Perception & Perspective Variance & \cite{CO_GG_DMPE, CO_GG_MVCL} \\ \cline{3-6}
& & 2D–3D & Cross-modality Enhancement & Sensor Registration & \cite{CO_GG_CoPeD} \\ \cline{2-6}

& \multirow{2}{*}{Air-Air} 
& 2D–2D & Long Baseline & Limited Bandwidth & \cite{CO_AA_UCDNet, CO_AA_MMUAV} \\ \cline{3-6}
& & 2D–3D & Distance-Aware Cooperation & Sensor Registration & \cite{CO_AA_Where2comm} \\ \cline{2-6}

& \multirow{2}{*}{Ground-Air} 
& 2D–2D & Cross-perspective Collab. & Limited View Alignment & \cite{CO_GA_GAC, CO_GA_ST} \\ \cline{3-6}
& & 2D–3D & Rich Information Support & Sensor Registration & \cite{CO_GA_TA} \\ \hline
\end{tabular}
}
\label{table:OB}
\end{table*}

\subsection{Low-level Feature Observation: Classic SLAM Front-end}
Low-level feature extraction, a cornerstone of classic SLAM front-ends, identifies geometric primitives like points, lines, and planes for foundational environmental observation.

\subsubsection{Point Feature}
Point features are fundamental geometric elements extracted and matched across frames. While classic gradient-based extractors like SIFT \cite{OB_P_SIFT} often fail in textureless scenes, deep learning has improved robustness. Learned extractors \cite{OB_P_SuperPoint} and transformer-based matchers \cite{OB_P_SuperGlue, OB_P_Loftr} now provide state-of-the-art performance for 2D images. Similar techniques have been developed for 3D point clouds from sensors like LiDAR \cite{OB_P_LOAM}.

\subsubsection{Structural Features}
In textureless environments, point features often provide insufficient information for reliable navigation. To address this limitation, structural features such as lines and planes are frequently employed. Classic methods leverage geometric principles, particularly the Manhattan world assumption, to detect line features \cite{OB_L_IVPR, OB_StructVIO}. More recently, end-to-end deep learning approaches have emerged for robust extraction and matching of line features \cite{OB_L_GlueStick}. Additionally, plane features extracted from depth information \cite{OB_P_STING} serve as effective complementary geometric primitives for environmental perception.


\subsection{High-level Cognitive Observation: Spatial Cognition}
Beyond low-level feature extraction, high-level cognitive observation focuses on semantic understanding of the environment through techniques such as object detection, semantic segmentation, and place recognition. This cognitive capability enhances the observation model by 

\begin{equation}
    O(\mathbf{o}_t|\mathbf{s}_t,\mathbf{a}_t) = \sum_{j=1}^n O(\mathbf{o}_t|\mathbf{s}_t,\mathbf{a}_t,c_j)p(c_j),
\end{equation}
where $c_j$ is the cognition variable, $p(c_j)$ is the probability of cognition $c_j$, and $n$ is the total number of possible cognitive states. $O(\mathbf{o}_t|\mathbf{s}_t,\mathbf{a}_t,c_j)$ represents the probabilistic observation model under cognition $c_j$. With high-level cognition, observations become more precise and informative. This cognitive enhancement is typically implemented through deep learning methods.

\subsubsection{Object Detection (OD)}
Object detection provides critical semantic information by localizing and classifying objects in 2D images and 3D point clouds. In 2D, the field has evolved from real-time CNNs (e.g., \cite{OB_OD_YOLOv5}) to transformers with enhanced contextual reasoning \cite{OB_OD_GCN}, while gaining robustness to environmental challenges like adverse weather \cite{OB_OD_Weather}. For 3D detection, methods address point cloud sparsity by fusing multimodal data \cite{OB_OD_ASIF_Net} or leveraging direct point geometry \cite{OB_OD_DCGNN}. This cognitive capability significantly enhances navigation by improving SLAM robustness through dynamic feature removal—implemented either directly \cite{OB_OD_DynaSLAM} or probabilistically \cite{OB_OD_Detect_SLAM}. Furthermore, OD enables semantically enriched mapping by focusing on specific objects \cite{OB_OD_RGBD} or establishing objects as persistent landmarks in object-level SLAM systems \cite{OB_OD_MonoOBSLAM, OB_OD_CubeSLAM}.

\subsubsection{Semantic Segmentation (SS)}
SS provides dense environmental understanding by assigning class labels to every pixel or point. In 2D, methods have evolved from foundational FCNs \cite{OB_SS_FCN} and context-aware architectures \cite{OB_SS_DeepLab} to more recent transformer-based approaches \cite{OB_SS_SegFormer}. For 3D point clouds, techniques range from pointwise MLPs \cite{OB_SS_Pointnet2} and specialized convolutions \cite{OB_SS_KPConv} to transformers \cite{OB_SS_PointTran}, with significant advances in foundation models \cite{OB_SS_Pointbert} and multimodal fusion \cite{OB_SS_MFFNet}. This semantic capability enhances environmental observation in several ways: facilitating dynamic object removal \cite{OB_SS_DS_SLAM}, strengthening geometric feature descriptions for robust mapping \cite{OB_SS_SuMa++}, and enabling SLAM systems to leverage semantic objects as reliable landmarks through probabilistic data association \cite{OB_SS_DA}. Additionally, semantic features can be implicitly integrated into end-to-end neural mapping networks to further improve observation quality \cite{OB_SS_SNI_SLAM}.

\subsubsection{Place Recognition (PR)}
Place Recognition serves as a critical component for loop closure detection and re-localization in navigation systems. In 2D applications, methods have evolved from CNN-based descriptors \cite{OB_PR_Netvlad} to advanced transformer architectures \cite{OB_PR_R2former} and universal foundation models \cite{OB_PR_Anyloc}. For 3D environments, LiDAR-based approaches span from classic descriptors \cite{OB_PR_OSK} to deep learning methods that effectively utilize temporal information \cite{OB_PR_SeqOT} or bird's-eye-view projections \cite{OB_PR_BEVPlace}. To enhance robustness across varied conditions, multimodal fusion techniques integrate 2D and 3D data through either implicit \cite{OB_PR_AdaFusion} or explicit \cite{OB_PR_EI} mechanisms. Practical implementation of these methods requires addressing system-level challenges including online calibration \cite{OB_PR_Cali} and ensuring real-time performance on robotic platforms \cite{OB_PR_UGV}.

\subsection{Multi-agent Observation: Collaborative Sensing}
Beyond single-agent perception, multi-agent collaborative sensing emerges as an effective strategy to expand perceptual coverage and enhance observation robustness. Heterogeneous observational perspectives—such as those from ground and aerial agents—provide complementary environmental information that can overcome individual sensor limitations and occlusions. This collaborative approach to environmental observation has been extensively studied within the field of multi-agent perception and has demonstrated significant improvements.

\subsubsection{Ground-Ground}
Ground-ground collaboration between multiple robotic agents significantly enhances perceptual coverage during navigation tasks. For efficient exploration and mapping, these systems employ techniques like dynamic Voronoi partitioning \cite{CO_GG_VB} or distributed potential fields \cite{CO_GG_DMPE} to effectively merge submaps and eliminate redundant work. To overcome communication bandwidth constraints, agents exchange compact data representations such as lightweight topological maps \cite{CO_GG_TOM, CO_GG_MR_TopoMap} or compressed descriptors \cite{CO_GG_SMMR} rather than transmitting dense map data. For robust decentralized localization, these collaborative systems can jointly learn and adapt to non-stationary sensor noise characteristics through variational Bayesian methods, thereby improving overall system accuracy and reliability \cite{CO_GG_DC, CO_GG_TV}.


\subsubsection{Air-Air}
UAVs' superior perceptive range and mobility make them particularly effective for collaborative observation tasks. Research in aerial multi-agent systems addresses several key challenges across different operational domains. In coordinated mapping and exploration, multi-UAV systems leverage collaborative SLAM techniques to enhance model accuracy \cite{CO_AA_MMUAV} while sharing submaps to achieve rapid environmental coverage \cite{CO_AA_RACER, CO_AA_Forest, CO_AA_UAV_LI}. Regarding advanced perception capabilities, researchers have developed frameworks for collaborative 3D object detection that fuse multi-view 2D images into coherent 3D representations \cite{CO_AA_UCDNet}. Additionally, to address the critical bottleneck of limited communication bandwidth, techniques such as spatial confidence maps have been implemented to intelligently compress and prioritize shared data between aerial agents \cite{CO_AA_Where2comm}.


\subsubsection{Ground-Air}
Ground-air collaborative perception integrates aerial and ground-based sensing perspectives to overcome the limitations of single-platform observation. Aerial agents typically perform rapid, coarse-grained mapping while ground agents capture detailed features in local regions \cite{CO_GA_AEM}, with the heterogeneous sensor data fused through global pose graph optimization techniques \cite{CO_GA_GAC}. The reliability of this sensor fusion process is further enhanced by incorporating high-level semantic features that bridge perspective differences \cite{CO_GA_ST}. Beyond general mapping, these collaborative systems can be optimized for specific objectives, such as task-oriented coordination for efficient target localization \cite{CO_GA_TA}. To facilitate systematic evaluation and advancement in this field, standardized benchmark datasets are established for these heterogeneous perception systems \cite{CO_GG_CoPeD}.

\subsection{Discussion and Trends}

The evolution of observation capabilities in EN systems reveals three key trajectories:

\begin{itemize}
    \item \textbf{From Static to Adaptive Sensing:} A shift from fixed parameters toward systems that dynamically adapt to environmental changes and sensor conditions.
    
    \item \textbf{From Passive to Active Perception:} A transition from passive data collection to strategic interaction for optimized information gathering.
    
    \item \textbf{From Predefined to Emergent Collaboration:} A progression from rigid protocols to emergent collaboration where agents learn optimal coordination strategies through interaction.
\end{itemize}


\section{Fusion of Proprioception and Exteroception: Sensing Fusion Intelligence}
The fusion stage integrates proprioceptive state transitions and exteroceptive observations to produce an optimal belief state. This state estimate is critical for the subsequent reward-policy construction. This section reviews two primary fusion paradigms: classic Bayesian methods and implicit neural techniques.

\begin{figure}
    \centering
\subfloat[Fusion pipeline and trend.]{\includegraphics[width=0.45\linewidth]{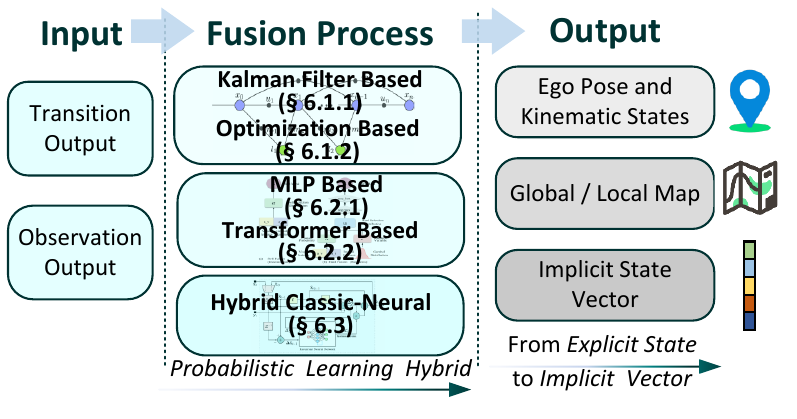}
    \label{fig:Structure}}
\subfloat[Reward-policy pipeline and trend.]{\includegraphics[width=0.58\linewidth]{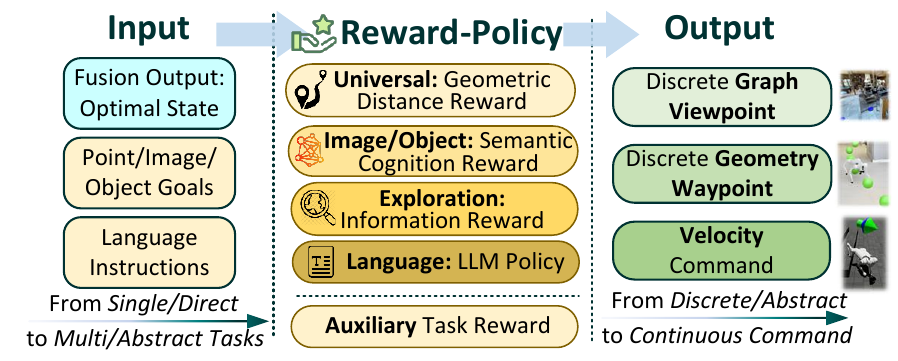}
    \label{fig:Structure}}
    \vspace{-0.3cm}
    \caption{A summary of the state fusion and reward-policy pipelines and trends.}
    \vspace{-0.5cm}
\end{figure}

\subsection{Classic Bayesian Based Fusion}
Classic fusion methods are grounded in Bayesian principles, whereby the optimal state is derived from integrating prior state beliefs with observation conditional probabilities according to
\begin{equation}
    b(\mathbf{s}_t) = \eta O(\mathbf{o}_t|\mathbf{s}_t,\mathbf{a}_t)\sum_{\mathbf{s}_t \in \mathcal{S}}T(\mathbf{s}_t|\mathbf{s}_{t-1},\mathbf{a}_t)b(\mathbf{s}_{t-1}),
\end{equation}
where $\eta$ is the normalization factor. Kalman filters and optimization-based approaches represent two predominant implementations of these classic state fusion methods.

\subsubsection{Kalman Filter Based Fusion}
Kalman filters represent a widely adopted approach for fusing proprioceptive and exteroceptive data in navigation systems. For visual-inertial fusion, the Multi-State Constraint Kalman Filter (MSCKF) serves as a classic framework implemented for both conventional cameras \cite{SP_Vector_OpenVINS} and asynchronous event cameras \cite{IF_KF_EVIO1}. Similarly, in range-inertial fusion, Kalman filters are effectively deployed across both well-established LiDAR-based systems \cite{IF_KF_FastLIO} and more challenging radar-based systems with inherently noisier measurements \cite{IF_KF_YRIO}. A fundamental challenge across all sensing modalities is maintaining robustness against non-Gaussian and non-stationary noise distributions, a problem addressed through specialized filtering techniques \cite{IF_KF_GGIG, IF_KF_ST, IF_KF_OR}. Contemporary state-of-the-art systems achieve enhanced robustness by tightly coupling multiple complementary modalities—including visual, LiDAR, and inertial data—within a unified filtering framework \cite{IF_KF_LIC, IF_KF_R3LIVE}.

\subsubsection{Optimization Based Fusion}
Optimization-based fusion methods construct a unified cost function for minimization that typically comprises prior costs, propagation costs, observation costs, and regularization terms. For vision-based systems, feature parameters are directly incorporated into the optimization process, as demonstrated in \cite{SP_Vector_VINS_Mono, IF_OP_ORBSLAM3, IF_OP_GRVIO}. Similarly, when working with event cameras, features extracted from asynchronous event streams are fused within the same optimization framework \cite{IF_OP_EVIO_CORN, IF_OP_PLEVIO, IF_OP_EIDV}. Range sensor data from LiDAR \cite{IF_OP_3DIO, IF_OP_LIO_SAM, IF_OP_ROSE} or mmWave Radar \cite{IF_OP_RIO} contribute structural features to the cost function. The strength of this approach lies in its ability to combine vision and range observations with proprioceptive measurements using covariance-based weighting schemes, yielding robust multi-modal fusion solutions \cite{IF_OP_LVI_SAM, IF_OP_LVIO_Fusion, IF_OP_LVI_AGR}.

\subsection{Neural Fusion}
In addition to Bayesian-based fusion methods, deep neural networks can be employed to implicitly fuse state transition and observation information.

\subsubsection{Multi-Layer Based Fusion}
Early neural fusion methods concatenate transition and observation features using multi-layer networks, a technique pioneered by VINet \cite{IF_CONCAT_VINET} and later extended to self-supervised frameworks \cite{IF_CONCAT_DeepVIO, IF_CONCAT_HVIOnet, IF_CONCAT_UNVIO}. However, simple concatenation is not robust to corrupted or misaligned sensor data. To improve robustness, selective fusion techniques were developed to identify and prioritize reliable features from various sensor modalities, thus mitigating the impact of sensor degradation \cite{IF_Seletive_SeleVIO, IF_Seletive_SelfVIO, IF_Selective_SSF}.

\subsubsection{Transformer Based Fusion}
Transformer architectures, with their attention mechanisms, offer a sophisticated approach for sensor fusion by establishing context-aware relationships between temporal sequences. Unlike simpler multi-layer methods, transformers can model complex dependencies across both spatial and temporal dimensions of heterogeneous sensor data \cite{IF_Attention_EMAVIO, IF_Attention_ATVIO}. The self-attention mechanism automatically assigns appropriate weights to measurements based on their reliability, effectively reducing the influence of noisy or corrupted inputs while preserving valuable information.

\subsection{Hybrid Classic-Neural Fusion}
Beyond pure neural approaches, hybrid classic-neural fusion strategies have emerged as a promising direction. This approach, proposed in \cite{HF_KalmanNet}, integrates classic Kalman filtering techniques with neural networks to effectively learn non-linear dynamics in the fusion process. The method has been successfully applied to GNSS/INS systems \cite{HF_KalmanNet_GINS} and shows significant potential for broader application within embodied sensing fusion frameworks. These hybrid approaches aim to combine the interpretability and theoretical guarantees of classic methods with the flexibility and learning capacity of neural networks.

\subsection{Discussion and Trends}
{The evolution of sensor fusion reveals three key trajectories:}

\begin{itemize}
    \item {\textbf{From \textbf{Fixed Parameter} to \textbf{Adaptive Fusion}:} Evolution from \textbf{static parameters} to \textbf{dynamic adjustment} for varying sensor conditions, now advancing toward \textbf{robust frameworks} that \textbf{generalize} across unseen data distributions.}
    
    \item {\textbf{From \textbf{State Estimation} to \textbf{Cognitive Generation}:} Shift from \textbf{low-dimensional} state regression toward generative approaches producing \textbf{high-dimensional} navigation cognition with \textbf{semantic understanding}.}
    
    \item {\textbf{From \textbf{Modality-Specific} to \textbf{Universal Fusion}:} Progression from \textbf{specialized sensor-specific} architectures toward \textbf{universal frameworks} incorporating \textbf{heterogeneous sensors} through learned \textbf{cross-modal representations}.}
\end{itemize}


\section{Reward-Policy Construction: Social Intelligence}
\label{sec:reward_policy}

The Reward-Policy Construction stage forms the agent's \textit{Social Intelligence} by translating high-level tasks into a reward function, $R(\mathbf{s}_t, \mathbf{a}_t)$, to guide policy learning. A typical navigation reward function is a weighted sum of several components:
\begin{equation}
\label{Eq: R(s,a)}
R_{\text{T}}(\mathbf{s}_t,\mathbf{a}_t) = w_0 I_{s} + w_1 D(\mathbf{s}_t,\mathbf{g}) + w_2 S(\mathbf{s}_t, \mathbf{a}_t) + w_3 E(\mathbf{a}_t),
\end{equation}
where the terms represent a sparse success reward ($I_s$), a dense progress reward ($D$), a safety penalty ($S$), and an effort cost ($E$). The agent learns a policy $\pi(\mathbf{s}_t)$ to maximize the expected cumulative reward. This section reviews reward-policy construction across three levels of increasing social complexity: single tasks, multiple tasks, and language-instructed tasks, reflecting the evolution of cognitive understanding and human-agent interaction.

\subsection{Single Task}
Single task navigation forms the foundational capability of EN systems. At this level, reward functions directly translate observable states into action guidance. Current research has categorized single-task navigation into four main types: point-goal navigation, image-goal navigation, object-goal navigation, and exploration.

\subsubsection{Point Goal}
In Point-Goal Navigation, agents navigate to target coordinates. While early methods used privileged GPS/compass data \cite{ST_PG_DDPPO}, recent work enables real-world deployment by integrating learned egocentric localization modules, allowing agents to navigate using only visual observations \cite{ST_PG_ELPG, ST_PG_VOEPG}.


\subsubsection{Image Goal}
In Image-Goal Navigation, the agent must reach the location where a reference image was captured by learning a policy that relates its current view to the goal, often via visual correspondence \cite{ST_IG_COR}. To this end, research has explored various policy architectures, including memory-augmented networks for long-horizon tasks \cite{ST_IG_MA}, modular designs for improved generalizability and explainability \cite{ST_IG_ML}, and transformer-based models that jointly process the sequence of observations, actions, and the goal to better capture temporal dependencies \cite{ST_IG_TR}. 


\subsubsection{Object Goal}
In Object-Goal Navigation, agents are tasked with locating objects of specified categories, a process that fundamentally relies on semantic scene understanding. A prevalent approach involves constructing explicit semantic maps to guide exploration strategies \cite{ST_OG_SE}, though researchers have also developed more efficient mapless methods that reduce computational overhead \cite{ST_OG_SSF}. System performance can be substantially enhanced by incorporating 3D object geometry, which provides more precise spatial understanding in complex environments \cite{ST_OG_3DSE}. Recent advanced methods extend beyond simple object detection to model commonsense relationships between objects, employing structured Graph Convolutional Networks \cite{ST_OG_HR} or more flexible transformer architectures \cite{ST_OG_VTNET} to optimize search efficiency. These approaches ultimately aim to develop high-level, abstract scene representations that facilitate more intelligent navigation behaviors \cite{ST_OG_AM, ST_OG_IM, ST_OG_GCEExp}.


\subsubsection{Exploration}
Exploration tasks focus on efficient coverage and reconstruction of unknown environments. \warning{The formulation of rewards varies with different representations of the agent's current state and explored areas}. Coverage-based approaches provide a straightforward reward mechanism based on the percentage of area the agent has successfully explored \cite{ST_EX_CV, ST_EX_CV_OA}. Complementing this strategy, intrinsic curiosity-driven rewards \cite{ST_EX_CURI} motivate the agent to seek novel information by prioritizing observations that differ significantly from previously encountered areas, thus promoting comprehensive environmental exploration.

\subsection{Multiple Tasks}
In multi-task navigation, research demonstrates that training agents to navigate toward one goal type can enhance performance across other navigation objectives \cite{MT_MI_VN, MT_MI_CA}. To address long-horizon tasks effectively, policies frequently incorporate persistent memory mechanisms, implemented either as explicit global semantic maps \cite{MT_MO_TPP, MT_MO_MultiON} or implicit learned representations \cite{MT_MO_NIR}, thereby minimizing redundant exploration. Navigation performance is further enhanced through auxiliary learning objectives, where agents simultaneously train on self-supervised tasks such as spatial reasoning \cite{MT_MO_SR} or predicting depth and inverse dynamics \cite{ST_PG_ATPGN}, resulting in more robust state representations. Although end-to-end deep learning approaches dominate recent research, classic modular architectures that decouple mapping from goal localization continue to demonstrate significant efficacy \cite{MT_MO_SGOLAM}.



\subsection{Visual Language Navigation}
In addition to the agent's self-directed policy learning, human language instructions can facilitate the learning process to accomplish complex tasks. This approach is referred to as visual language navigation (VLN). The language instructions are typically encoded and integrated into the policy model to guide the agent, while the underlying reward formulation remains unchanged. A straightforward approach involves encoding language as an input feature to the policy model, as detailed in \cite{LI_TFS_VLGN}. With the advancement of large language models (LLMs), natural language can be processed through these models for enhanced comprehension. One approach incorporates LLMs for zero-shot navigation tasks; in \cite{LI_PFT_LLM}, the authors utilize LLMs as zero-shot navigation code generators for policy construction. Alternatively, without fine-tuning the LLM itself, retrieval-augmented generation (RAG) techniques have been introduced into the EN process \cite{EN_RAG,NavRag}. These RAG-based methods leverage the logical reasoning capabilities of LLMs to process specific navigation graph memories. For navigation foundation models, researchers in \cite{LI_PFT_BEVBert, LI_PFT_VLM, LI_PFT_ITP, LI_PFT_PRE} have pre-trained and fine-tuned specialized navigation LLMs to process language instructions, enabling agents to generate appropriate navigation policies.

\subsection{{Discussion and Trends}}
{The evolution of reward-policy construction reveals a progression toward more flexible and intuitive human-agent interaction:}

\begin{itemize}
    \item {\textbf{From Explicit to Implicit Goals:} Evolution from \textbf{geometric coordinates} to \textbf{semantic concepts}, enabling more abstract reward functions.}
    
    \item {\textbf{From Single to Multi-Task Learning:} Progression from \textbf{specialized solvers} to \textbf{generalized policies} that transfer knowledge across tasks.}
    
    \item {\textbf{From Predefined Rewards to Language Interpretation:} Shift toward deriving rewards from \textbf{natural language instructions} via \textbf{language models}, replacing predefined functions.}
    
    \item {\textbf{From Discrete to Differentiable Rewards:} Transition from \textbf{sparse, binary signals} to \textbf{dense, continuous formulations} with smooth gradients for more efficient policy optimization.}
\end{itemize}


\section{Action Execution: Motion Intelligence}
Following reward-policy construction, an agent must employ specialized motion skills to effectively implement the action sequence generated by the policy.
Following reward-policy construction, the action execution stage represents the motion intelligence. This section systematically categorizes action capabilities across three levels of increasing complexity: meta skills, combination skills, and morphological collaboration.

\subsection{Meta Skills Development}
Meta skills are foundational action primitives for basic mobility. For wheeled robots, these include omnidirectional movement and obstacle avoidance, often learned effectively through unsupervised methods \cite{AS_SS_W} due to simpler dynamics. Legged robots, with their high-dimensional action spaces, require more complex approaches. Research has focused on model-based control for precise maneuvers like jumping \cite{AS_SS_JP} and learning-based techniques that use reinforcement learning, often with curriculum learning \cite{AS_SS_CL_Gait}, to discover stable gaits through environmental interaction \cite{AS_SS_RL+, AS_SS_LG, AS_CB_HL}.

\subsection{Combination Skills}
Combination skills integrate multiple meta skills into coherent, higher-level capabilities—representing a critical advancement in motion intelligence. 

\subsubsection{Sequential Skill Composition}
Sequential skill composition enables legged robots to perform complex maneuvers. Explicit, hierarchical approaches \cite{AS_CB_HRL, AS_CB_JC} use high-level planners to sequence skills, offering interpretability but potentially lacking smooth transitions. In contrast, implicit, end-to-end learning methods \cite{AS_CB_Parkour, AS_CB_WOCOCO} create unified policies for seamless skill transitions, demonstrating impressive parkour-like agility at the cost of extensive training.

\subsubsection{Leg-Arm Manipulation}
Beyond pure locomotion, recent research has explored dual-purpose capabilities where legged systems simultaneously handle locomotion and manipulation. Leg-based manipulation \cite{AS_CB_Pedipulate} enables object interaction without dedicated end-effectors, while integrated arm-leg systems \cite{AS_CB_UMI, AS_CB_HiLMa} support sophisticated whole-body coordination. These systems implement complex motion planning algorithms that manage the coupled dynamics between locomotion and manipulation subsystems. 

\subsection{Morphological Collaboration}
Morphological collaboration represents the frontier of motion intelligence, where robots leverage adaptable physical configurations to optimize performance across diverse environments.

\subsubsection{Wheel-Legged Hybrids}
Wheel-legged robots exemplify this approach, combining wheeled efficiency on flat terrain with legged versatility on irregular surfaces. These hybrid platforms present unique control challenges addressed through specialized frameworks \cite{AS_MC_WL, AS_MC_WL_MPC} that dynamically allocate control authority between wheel and leg subsystems based on terrain classification. 

\subsubsection{Terrestrial-Aerial Robots}
Terrestrial-aerial robots \cite{AS_MC_SytaB, AS_MC_TAB} represent another class of morphologically collaborative systems, capable of ground locomotion and aerial flight. These platforms implement sophisticated state estimation algorithms that maintain localization coherence across drastically different dynamics regimes, while specialized motion planners manage the complex transition phases between locomotion modes. Energy optimization becomes particularly critical for these systems, with research focusing on trajectory optimization techniques that minimize power consumption across mode transitions \cite{AS_MC_SytaB}.

\subsubsection{Aerial-Aquatic Robots}
The most advanced morphological collaborations extend to multi-domain robots capable of operating across aerial, terrestrial, and aquatic environments \cite{AS_MC_AA}. These systems implement distributed control architectures where specialized controllers for each domain are coordinated through a meta-control layer. Despite their promise, these sophisticated systems face significant challenges in reliability, energy efficiency, and control complexity—areas.

\subsection{{Discussion and Trends}}
{The evolution of motion execution approaches reveals three key transformations:}

\begin{itemize}
    \item {\textbf{From Scene-Specific to Morphological Adaptation:} Evolution from \textbf{environment-specialized} systems toward agents that \textbf{dynamically reconfigure} themselves for diverse, unpredictable terrains.}
    
    \item {\textbf{From Precision Control to Task-Oriented Execution:} Shift from \textbf{perfect trajectory following} toward \textbf{pragmatic goal achievement} despite imperfect control, especially evident in TN systems.}
    
    \item {\textbf{From Predefined to Interactive Motion:} Development of capabilities that allow agents to refine movement strategies through \textbf{environmental interaction} rather than relying solely on \textbf{pre-programmed patterns}.}
\end{itemize}

\section{Systematic Embodied Navigation Approaches}
Systematic EN approaches integrate the entire TOFRA pipeline. These range from classic modular systems with explicit interfaces \cite{ClassicExploration} to end-to-end learning frameworks where TOFRA stages are implicitly connected through neural network layers.
\warning{This section introduces systematic EN approaches applied across three representative navigation scenarios.}






\begin{table*}
\centering
\caption{Embodied Navigation Systems and References}
\renewcommand{\arraystretch}{1.5} 
\setlength{\tabcolsep}{0.1cm} 
\resizebox{1.0 \textwidth}{!}{
\begin{tabular}{c|c|c|c|c|c}
\hline
\textbf{Scene} & \textbf{Input End} & \textbf{Output End} & \textbf{Properties} & \textbf{Challenges} & \textbf{References} \\ \hline

\multirow{6}{*}{AD} 
& GPS/Map + Surrounding Cameras & Control & Low-cost, First Principle & High-speed, weather conditions & \cite{EE_FRA_SRC} \\ \cline{2-6}
& GPS/Topo Map + Image & Control + Localization & Geospatial awareness & Map updates & \cite{EE_FR_VNN} \\ \cline{2-6}
& GPS/Topo Map + LiDAR & Control + Uncertainty & Robust distance sensing & Computational cost & \cite{EE_FR_LN} \\ \cline{2-6}
& GPS/Map + Camera/LiDAR/Radar & Control & Robustness to various conditions & Computational cost & \cite{EE_FRA_PMS} \\ \cline{2-6}
& Images + Instructions & Waypoints + Control & Task-driven navigation & Instruction interpretation, visual dependency & \cite{EE_TOFR_WP} \\ \cline{2-6}
& Semantic Images + Instructions & Control & Contextual understanding & Semantic inconsistency & \cite{EE_TOFRA_Pre} \\ \hline

\multirow{5}{*}{IN} 
& Observational Images & Control & Visual input, real-time control & Limited field of view, light conditions & \cite{EE_TOFR_MP} \\ \cline{2-6}
& Observational/Goal Images & Control + Map & Goal-driven navigation & Path planning complexity & \cite{EE_TOFR_VN} \\ \cline{2-6}
& SE(2) State + Image & Control + Semantic & State estimation and visual fusion & Ambiguity in semantics & \cite{EE_FRA_XMob} \\ \cline{2-6}
& Motion + LiDAR & Control + Trajectory & Real-time motion tracking & Trajectory prediction & \cite{EE_TOFRA_NeuPAN} \\ \cline{2-6}
& NeRF Memory + Image & Control & Neural network-based memory & Computational cost & \cite{EE_TOFRA_NerfNav} \\ \hline

\multirow{2}{*}{TN} 
& Joints + Depth & Joint Control & Direct control of joints & Depth Noise & \cite{EE_TOFRA_MN} \\ \cline{2-6}
& Joints + Corrupted Depth & Velocity Control & Incomplete sensory data & Error propagation & \cite{EE_TOFR_CP} \\ \hline

\end{tabular}
}
\begin{tablenotes}
    \scriptsize
    \item Note: AD refers to Autonomous Driving, IN refers to Indoor Navigation, and TN refers to Terrain Navigation.
\end{tablenotes}
\label{table:EE}
\vspace{-0.6cm}
\end{table*}

\subsection{Embodied Autonomous Driving (AD) System}
\warning{
In Embodied Autonomous Driving (AD), systems leverage external GPS and maps, effectively treating the state transition (T) as a given. The research focus is thus on learning a robust end-to-end policy for the observation, fusion, reward-policy, and action (OFRA) stages. These policies learn to map increasingly rich perceptual inputs—from surround-view cameras \cite{EE_FRA_SRC}, to 3D LiDAR \cite{EE_FR_LN}, and weather-resilient radar \cite{EE_FRA_PMS} to vehicle control. Policy sophistication has also advanced, from producing probabilistic outputs to handle uncertainty \cite{EE_FR_VNN} to incorporating human guidance for fine-tuning \cite{EE_TOFR_WP, EE_TOFRA_Pre}. Ultimately, these systems implicitly model a complex OFRA pipeline on top of an explicit state transition.
}
\subsection{Embodied Indoor Navigation (IN) System}
Unlike autonomous driving systems, Embodied Indoor Navigation (IN) systems operate without GPS, relying entirely on egocentric sensors. These systems vary in complexity across the TOFRA pipeline. Policies directly mapping observations to immediate actions are effective for short-horizon tasks but limited in planning capability \cite{EE_TOFR_MP, EE_TOFR_VN}. More advanced approaches incorporate explicit state representations alongside learned transition models implemented through recurrent networks or transformers \cite{EE_FRA_XMob, EE_T_VO}. To enhance spatial understanding, sophisticated IN systems build rich world models using either semantic features \cite{EE_FRA_XMob} or Neural Radiance Fields (NeRF) \cite{EE_TOFRA_NerfNav}. These world models significantly improve the agent's environmental comprehension and navigation decision-making in complex indoor settings.

\subsection{Embodied Terrain Navigation (TN) System}
\warning{
Terrain Navigation (TN) focuses on traversing complex, uneven landscapes to accomplish navigation tasks. These systems typically employ legged robots as their primary agent platforms. The input typically consists of joint sensor data from the agent and surrounding depth maps of the terrain. For output, TN systems either generate velocity commands for low-level controllers to track \cite{EE_TOFR_CP} or provide direct joint control commands \cite{EE_TOFRA_MN}. While velocity-based approaches may suffer from discontinuities in motion execution, direct joint control offers smoother operation but presents greater challenges for policy learning.
}

\subsection{{Discussion and Trends}}
The evolution of systematic EN approaches reveals three key transformations:

\begin{itemize}
    \item {\textbf{From Specialized to Adaptive:} Evolution from \textbf{scene-specific} systems toward \textbf{unified frameworks} that adapt dynamically to diverse environments without retraining.}
    
    \item {\textbf{From Modular to Integrated:} Shift from optimizing individual TOFRA components separately toward \textbf{joint optimization} strategies that address cross-system dependencies.}
    
    \item {\textbf{From Engineered Task to Emergent Intelligence:} Progression from \textbf{human-designed} navigation hierarchies toward systems that \textbf{autonomously decompose} complex instructions through \textbf{continual learning}.}
\end{itemize}

\section{Open Platforms and Evaluation Metrics}
This section reviews platforms and metrics for EN evaluation. We analyze key simulators and real-world platforms through the TOFRA framework and discuss metrics for assessing social, sensing, and motion intelligence.

\begin{figure*}
    \centering
    \includegraphics[width=0.7\linewidth]{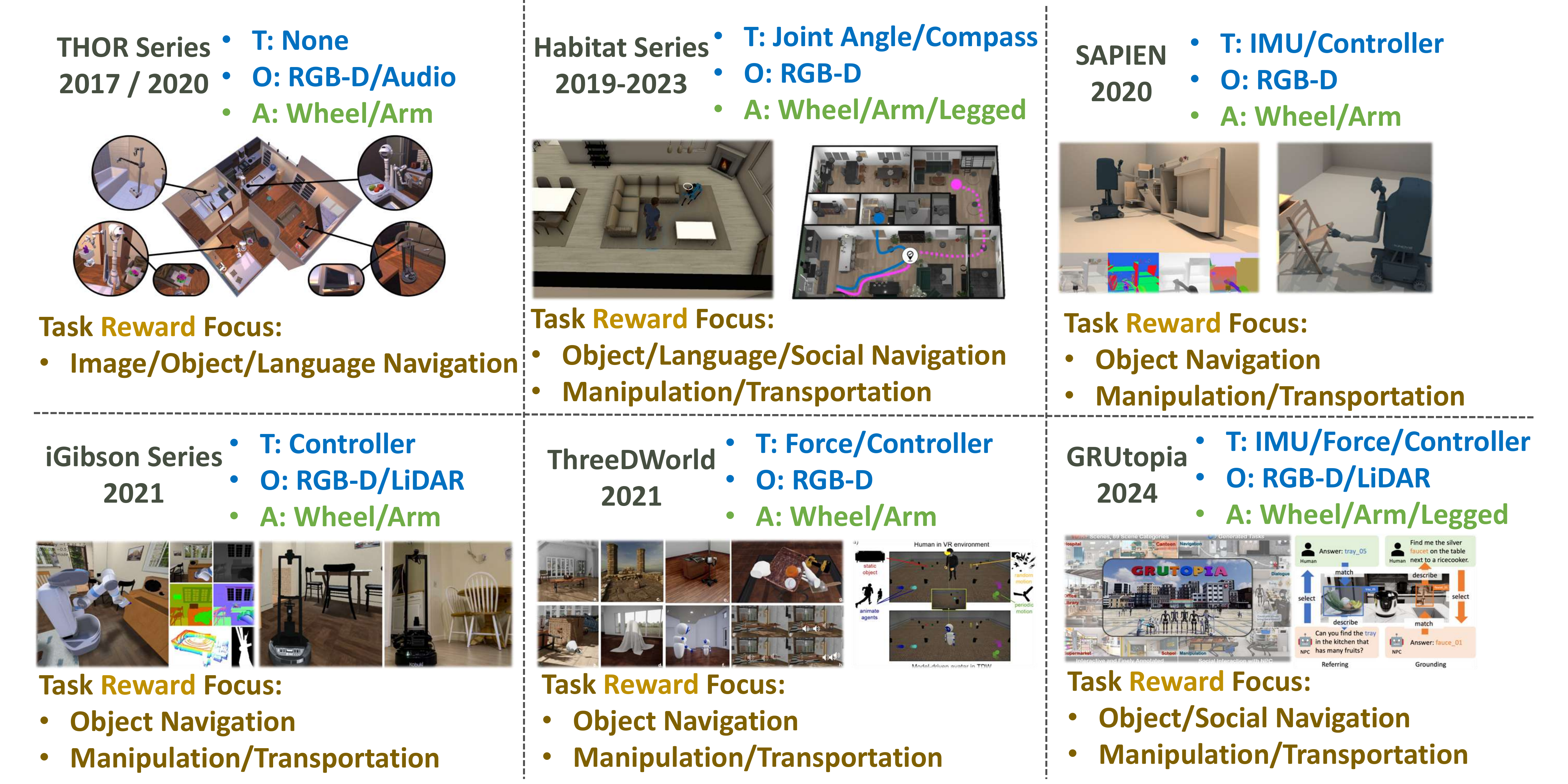}
    \caption{Fundamental EN simulators and supporting modules within the TOFRA framework.}
    \label{fig:ENPlatform}
    \vspace{-0.5cm}
\end{figure*}

\subsection{Platforms}

\subsubsection{Simulators}
Simulators provide scalable, safe, and reproducible environments for developing and benchmarking EN systems. Key platforms include:
\begin{itemize}
    \item \textbf{AI2-THOR \cite{PD_SIM_AI2THOR}:} An interactive, room-scale simulator focused on social/language tasks (R) and manipulation (A). Its companion, RoboTHOR \cite{PD_SIM2REAL_RoboTHOR}, facilitates direct sim-to-real transfer.
    \item \textbf{Habitat Series \cite{PD_SIM_Habitat, PD_SIM_Habitat2, PD_SIM_Habitat3}:} A highly optimized platform that has progressively evolved to support complex, building-scale navigation. It now incorporates manipulation, legged robots, and language instructions, expanding its coverage across the entire TOFRA framework.
    \item \textbf{SAPIEN \cite{PD_SIM_SAPIEN} and iGibson \cite{PD_SIM_iGibson}:} These simulators enhance physical realism by providing detailed proprioceptive sensors (T) like IMUs (SAPIEN) and exteroceptive sensors like LiDAR (iGibson).
    \item \textbf{ThreeDWorld \cite{PD_SIM_3DWorld}:} Extends simulation to outdoor scenes and includes force feedback for more realistic physical interaction.
    \item \textbf{GRUtopia \cite{GRUtopia}:} Built on Issac Sim \cite{PL_Orbit}, this platform offers high photo- and physical-realism. It supports the full TOFRA pipeline for sensing, social, and motion intelligence, though it is computationally demanding.
\end{itemize}

\subsubsection{Real-world Platforms}
While simulators are essential for development, validation on physical hardware is crucial. Standard wheeled platforms include the TurtleBot Series, Fetch and Freight, and Clearpath Robotics (Jackal, Husky). For complex terrain navigation, common legged platforms are the Boston Dynamics Spot, ANYbotics ANYmal, and Unitree Series (Go1, A1).

\begin{figure*}
    \centering
    \includegraphics[width=0.8\linewidth]{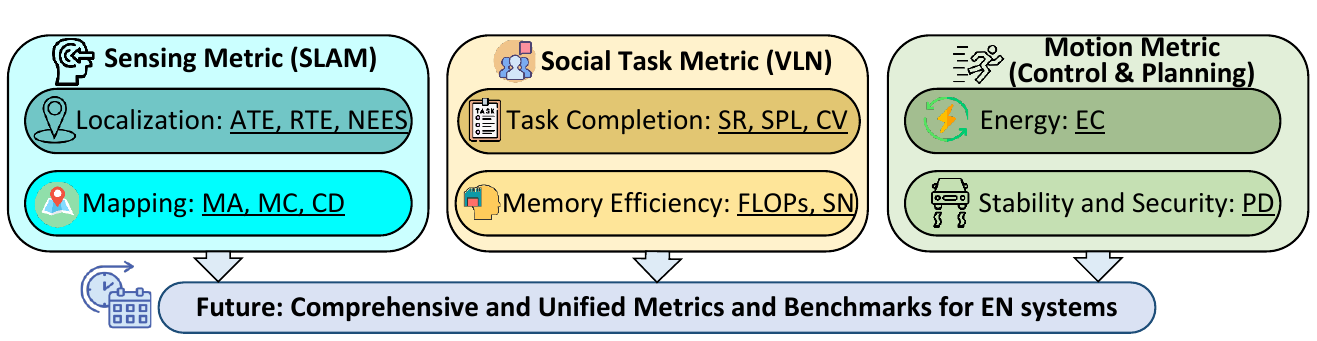}
    \caption{An overview of existing major metrics for embodied navigation system.}
    \vspace{-0.5cm}
    \label{fig:ENEval}
\end{figure*}
\subsection{Evaluation Metrics}





A comprehensive evaluation of an EN system requires assessing its performance across three fundamental dimensions aligned with the TOFRA framework: social, sensing, and motion intelligence.

\subsubsection{Social Task Level Metrics}
Social task performance is quantified through several established metrics. For goal-oriented navigation, Success Rate (SR) measures the percentage of successfully completed tasks, while Success weighted by Path Length (SPL) evaluates efficiency by penalizing suboptimal paths \cite{LI_TFS_VLGN}. Computational efficiency is assessed through Floating Point Operations (FLOPs), quantifying the policy model's resource requirements \cite{EV_FLOPs}. For exploration-based tasks, system performance is characterized by Coverage (CV), which measures the proportion of environment explored, and Storage Node count (SN), which evaluates memory efficiency of the resulting spatial representation \cite{BIG}.

\subsubsection{Sensing Level Metrics}
Sensing capabilities are assessed through metrics targeting both localization and mapping functions. For localization performance, Absolute Trajectory Error (ATE) quantifies global positioning consistency, while Relative Trajectory Error (RTE) measures short-term odometry drift \cite{EV_ATE}. The statistical robustness of probabilistic estimators is evaluated using Normalized Estimation Error Squared (NEES), which examines the consistency between estimated uncertainties and actual errors \cite{EV_NEES}. Regarding mapping performance, Map Accuracy (MA) evaluates geometric precision \cite{EV_MA}, Map Completion (MC) determines environmental coverage completeness \cite{EV_MC}, and Chamfer Distance (CD) provides a quantitative measure of overall reconstruction fidelity between generated and reference point clouds \cite{EV_CD}.

\subsubsection{Motion Level Metrics} 
Motion execution metrics assess how effectively an agent implements its navigation policy through physical movement. Two key metrics quantify this performance: Energy Consumption (EC), which measures the total power expended during navigation tasks\cite{EV_EC}, and Path Deviation (PD), which evaluates stability by measuring how closely the agent adheres to its intended trajectory\cite{EV_PD}.

\subsection{{Discussion and Trends}}
{The evolution of EN platforms and metrics reflects a push toward greater realism and more comprehensive evaluation, aligning with the holistic TOFRA framework.}

\begin{itemize}
    \item {\textbf{Platforms Evolution:} A shift from \textbf{abstract, task-focused simulators} to \textbf{high-fidelity 'digital twins'} with realistic physics, with an emerging focus on \textbf{fully differentiable simulation environments} that enable \textbf{end-to-end gradient-based optimization} across the entire TOFRA pipeline, facilitating \textbf{joint training} of \textbf{perception, planning, and control} components.}
    
    \item {\textbf{Metrics Trend:} An expansion from \textbf{singular task metrics} (e.g., SR) to a \textbf{'balanced scorecard'} that \textbf{holistically assesses} an agent's \textbf{Social (R), Sensing (T,O,F), and Motion (A) intelligence}.}
\end{itemize}


\begin{figure*}
    \centering
    \includegraphics[width=0.8\linewidth]{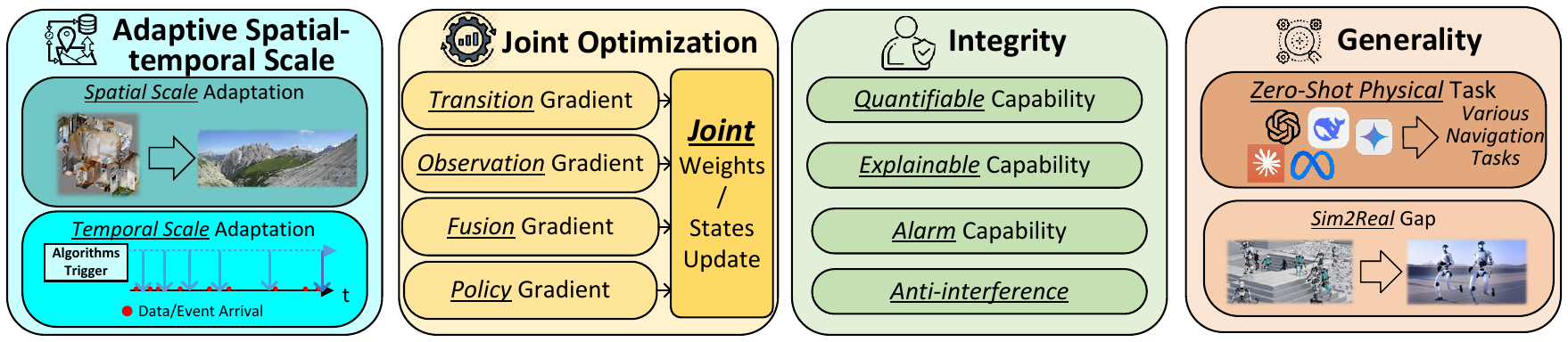}
    \caption{Major challenges and open research problems of EN systems.}
    \label{fig:ENEval}
    \vspace{-0.5cm}
\end{figure*}

\section{Open Research Problems}

Despite recent progress, achieving fully autonomous EN in real-world scenarios remains a significant challenge. This section identifies key open research problems and discusses future directions.

\subsection{Adaptive Spatial-temporal Scale for the EN system}
\warning{The operating spatial-temporal condition of an EN agent is crucial for its effective application, as it is influenced by the agent's observation capabilities, fusion strategies, and task cognition skills. To improve robustness and long-term autonomous task performance, advanced adaptive spatial-temporal scale strategies is essential.}

\subsubsection{Adaptive Spatial Scale}
EN systems often fail to generalize across different spatial scales. Classic methods are limited by scale-dependent hyperparameters in their optimization processes \cite{BioPR}, while end-to-end learning approaches struggle with feature distribution shifts that invalidate offline-trained weights \cite{IMOST,RobustFlight}. A key future direction is developing incremental update mechanisms for both hyperparameters and network weights to enable seamless adaptation to spatially-varying scenes.

\subsubsection{Adaptive Temporal Scale} 
Adaptive temporal scaling is essential for managing high-frequency sensor data on resource-constrained agents. EN systems must learn to schedule T/O/F/R/A computations efficiently \cite{THE-SEAN}, which involves optimizing task selection ($\mathbf{u}_t \in \{T, O, F, R, A\}$) to minimize performance loss under computational constraints $\mathcal{C}$:
\begin{equation}
\warning{\min _{\mathcal{T}} \sum_{t=1}^T\mathcal{L}_{\text {task }}(t) \quad \text { s.t. } \quad \mathbf{u}_t=\mathcal{T}\left(\mathbf{x}_t, \mathcal{C}, t\right).}
\end{equation}
\warning{Future work should focus on developing efficient schedulers that trigger EN modules based on data relevance and task priority.}




\subsection{Joint Optimization of the EN system}
\warning{Another key challenge of EN is the joint optimization of all EN system parameters ($T, O, F, \pi$) by minimizing a total loss:}
\begin{equation}
    \warning{\min _{\{T, O, F, \pi\}} \sum_{t=1}^T \mathcal{L}_{\text {total }}(T, O, F, \pi). }
\end{equation}
\warning{Classic modular methods optimize components like SLAM and RL independently, ignoring interdependencies. Conversely, end-to-end learning enables joint optimization but suffers from poor generalization and high data requirements. Future work should focus on hybrid methods that integrate the strengths of both classic and learning-based approaches for robust, jointly optimized EN systems.}



\subsection{Integrity of the EN system}
Integrity, crucial for traditional navigation, is even more complex in EN due to dynamic environments and perception \cite{EN_Risk, EN_Secure}. It ensures reliable and safe operation by addressing key capabilities:
\begin{itemize}
    \item \textbf{Quantifiable Capability:} The ability to statistically measure and report confidence in its navigation outputs.
    \item \textbf{Explainable Capability:} The capacity to provide human-interpretable rationale for its decisions, potentially by mapping neural network states to understandable explanations.
    \item \textbf{Alarm Capability:} The ability to detect and alert for anomalous or unsafe conditions when system integrity is compromised.
    \item \textbf{Anti-interference Capability:} The robustness to maintain performance against external disturbances, such as sensor corruption or adversarial actions.
\end{itemize}



\subsection{Generality of Data and Task}
Generalizing across diverse data and tasks remains a primary challenge for deep learning-based EN. While some systems show zero-shot capabilities \cite{ST_IG_ZSON}, their success is often confined to small-scale, controlled environments. The sim-to-real gap presents a significant barrier, as the large-scale nature of navigation amplifies discrepancies between simulated and real-world data. Bridging this gap requires either physical photorealistic simulators or teaching agents abstract, transferable knowledge. Consequently, a unified benchmark is crucial for systematically evaluating the generalization capabilities of EN systems.




\section{Conclusion}
This survey introduced the TOFRA framework to unify EN research, revealing a synthesis of classic robotics and data-driven AI for navigation. Key trends include a shift from fixed models to \textbf{adaptive evolving intelligence}, from specialized components to \textbf{integrated systems}, and from engineered solutions to \textbf{emergent, language-guided behavior}. Despite this progress, major challenges persist in developing principled hybrid adaptive models, bridging the sim-to-real gap, and achieving long-horizon reasoning and navigation. We hope this survey encourages integrated approaches toward creating truly autonomous agents.


\bibliographystyle{ACM-Reference-Format}
\bibliography{ref_suv}

\end{document}